\documentclass{article} 
\usepackage{iclr2027_conference,times}

\usepackage{amsmath,amsfonts,bm}

\def\eqref#1{equation~\ref{#1}}

\def\1{\bm{1}}

\DeclareMathAlphabet{\mathsfit}{\encodingdefault}{\sfdefault}{m}{sl}
\SetMathAlphabet{\mathsfit}{bold}{\encodingdefault}{\sfdefault}{bx}{n}

\usepackage{hyperref}
\usepackage{url}
\usepackage{graphicx}
\usepackage{xcolor}
\definecolor{TableBest}{HTML}{FFF2A8}
\newcommand{\bestvalue}[1]{{\setlength{\fboxsep}{0.8pt}\colorbox{TableBest}{#1}}}
\usepackage{booktabs,array,multirow}
\usepackage{placeins}
\usepackage{float}
\usepackage{capt-of}
\usepackage{caption}

\title{DexWeave: Learning Dexterous Humanoid Loco-Manipulation from Human Demonstrations}

\author{
\textbf{Naichuan Sun}\textsuperscript{1,2}\thanks{Equal contribution.} \quad
\textbf{Haotian Shen}\textsuperscript{1}\footnotemark[1] \quad
\textbf{Yizhang Zhang}\textsuperscript{1} \quad
\textbf{Luying Feng}\textsuperscript{1} \quad 
\textbf{Haoze Wang}\textsuperscript{1} \quad \\
\hspace{2.5pt}\textbf{Yuanbo Xiangli}\textsuperscript{2} \quad 
\textbf{Yaochu Jin}\textsuperscript{1} \quad
\textbf{Peidong Liu}\textsuperscript{1}\thanks{Corresponding author.} \\
\textsuperscript{1}Department of Artificial Intelligence, School of Engineering, Westlake University \\
\textsuperscript{2}School of Artificial Intelligence, Shanghai Jiao Tong University \\
\texttt{naichuan.sun@sjtu.edu.cn} \\
\texttt{\{shenhaotian,liupeidong\}@westlake.edu.cn} \\
}

\iclrfinalcopy 
\begin{document}
\maketitle
\vspace{-5mm}
\noindent\begin{minipage}{\textwidth}
\centering
\includegraphics[width=\linewidth]{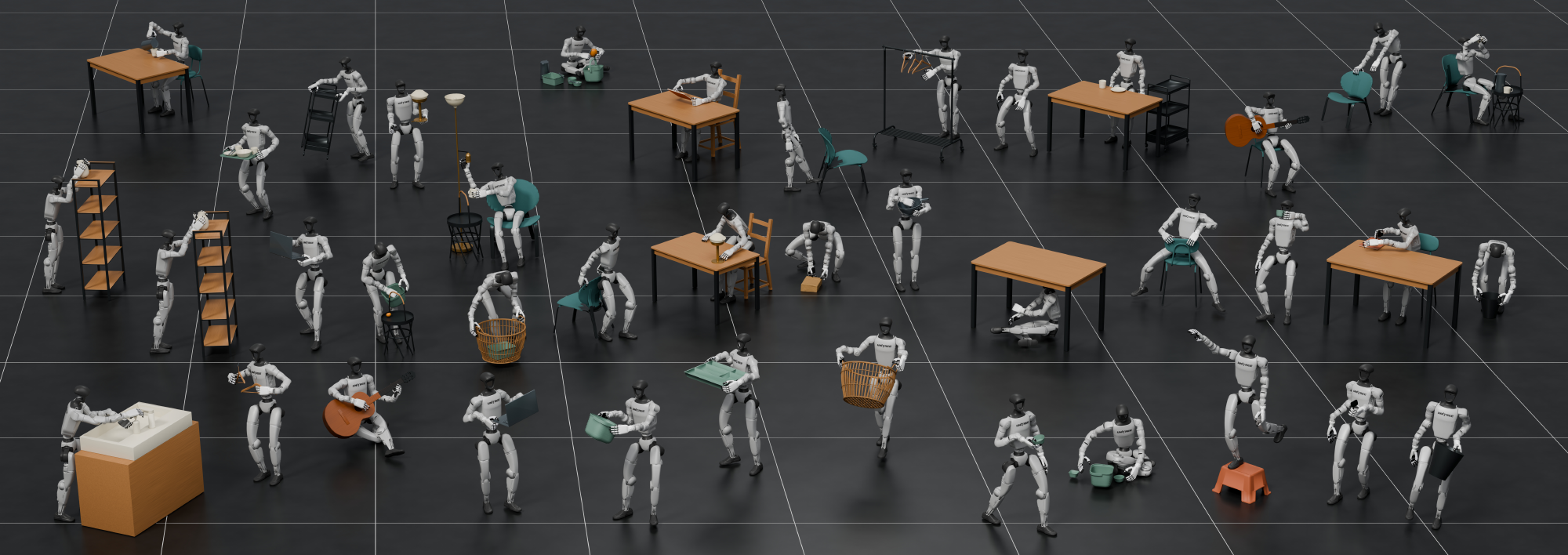}
\captionsetup{skip=3pt}
\captionof{figure}{\textbf{DexWeave} translates human demonstrations into executable dexterous humanoid loco-manipulation skills.}
\label{fig:teaser}
\end{minipage}
\vspace{2mm}

\begin{abstract}

Learning dexterous humanoid loco-manipulation from human demonstrations requires transferring not only human motion, but also the coordinated interaction structure underlying the demonstrated behavior. This is challenging because embodiment differences distort the coupling among body motion, wrist placement, finger articulation, and object interaction, while kinematically accurate references may still be difficult to realize under robot dynamics. We present \textbf{DexWeave}, a unified framework that connects interaction-consistent motion retargeting with anatomy-aware whole-body policy learning. DexWeave first employs a two-stage retargeting procedure that initializes body and hand motions with specialized solvers and subsequently performs coupled refinement over the upper-body interaction chain while preserving lower-body support. The resulting references are tracked by an anatomy-aware Transformer policy that represents anatomical regions as structured tokens and uses directed masked attention to model their dependencies, with object information selectively conditioning the upper-body pathway for dexterous interaction. The policy jointly outputs body and dexterous-hand actions and is trained directly with reinforcement learning, without pretrained tracking policies, teacher--student distillation, or subsequent residual refinement. DexWeave improves retargeting fidelity and interaction consistency while achieving higher manipulation performance and faster policy convergence than MLP baselines. We further deploy the learned policies on a physical Unitree G1 humanoid equipped with Inspire dexterous hands, demonstrating dexterous whole-body loco-manipulation in the real world. See our \href{https://dexweave.github.io}{project page} for videos.

\end{abstract}

\section{Introduction}\label{sec:introduction}

Dexterous humanoid loco-manipulation could enable a wide range of real-world tasks, but requires tight coordination between locomotion, whole-body support, and articulated hand--object interaction. Learning from demonstrations has emerged as a scalable approach to acquiring robot manipulation and humanoid whole-body skills~\citep{chi2023diffusionpolicy,zitkovich2023rt2,kim2025openvla,allshire2025videomimic,heng2026humdex}. In particular, abundant human motion and video data provide rich demonstrations of coordinated whole-body behaviors without requiring costly robot teleoperation~\citep{he2024omnih2o,heng2026humdex,allshire2025videomimic,jiang2026wholebodyvla}. Leveraging such data for dexterous humanoids, however, requires retargeting both body motion and hand articulation to the robot's morphology while preserving the demonstrated interactions, together with a control policy that can realize these interactions under robot dynamics.


A common challenge underlies both retargeting and control: wrist placement and finger articulation jointly determine hand--object interaction, yet they are often optimized at different levels of the motion hierarchy. Whole-body retargeting approaches improve motion fidelity and kinematic feasibility, while dexterous-hand methods primarily optimize hand-level kinematics and local hand--object geometry~\citep{handa2020dexpilot,xin2025objectives,wu2026toporetarget}. Despite advances in teleoperation and physics-based refinement~\citep{heng2026humdex,pan2025spider}, these levels can remain weakly coupled: differences in arm reach, palm geometry, and finger dimensions may make wrist placement and finger articulation mutually inconsistent even when each is locally well retargeted. The same coupling reappears during control: manipulation forces perturb whole-body balance, while locomotion continuously changes wrist placement and can disrupt hand--object contact. Existing policies address whole-body coordination through jointly trained upper- and lower-body agents~\citep{zhang2026falcon}, unified policies for locomotion and upper-body control~\citep{sun2025ulc}, or pretrained body and hand priors coordinated through a residual policy~\citep{li2026coordex}. While these approaches make high-dimensional control more tractable, fine-grained coordination between whole-body motion and articulated hand interaction remains challenging. Our key insight is that successful human-to-humanoid dexterous skill transfer requires preserving demonstrated interaction structure across embodiments and exploiting anatomical structure as an inductive bias for whole-body policy learning.

Guided by this insight, we present \textbf{DexWeave}, an integrated framework that bridges interaction-consistent motion retargeting and anatomy-aware whole-body policy learning. A two-stage retargeting pipeline first initializes robot motion using specialized body and hand solvers with contact-aware wrist blending, and then jointly refines the arms, wrists, and fingers along the upper-body interaction chain to preserve wrist--finger--object coordination while maintaining feasible lower-body support. An anatomy-aware Transformer organizes whole-body control through regional body tokens and directed masked attention that captures structured dependencies across anatomical regions. Object information is selectively injected into the upper-body pathway, conditioning dexterous interaction while leaving the lower-body pathway unconditioned on direct object observations. For each reference motion, a single policy network jointly outputs body and dexterous-hand actions and is trained with Proximal Policy Optimization (PPO), without tracking-policy pretraining, teacher--student distillation, or subsequent residual refinement.

We evaluate DexWeave primarily on whole-body interactions involving fine-grained dexterous hand--object contact from GRAB~\citep{taheri2020grab} and HUMOTO~\citep{lu2025humoto}, with additional evaluations on whole-body object interaction from OMOMO~\citep{li2023object} and body-only motion from LAFAN1~\citep{harvey2020robust}. Retargeting comparisons show improved kinematic feasibility, with reduced penetration and foot skating, while more accurately preserving hand--object interactions. For closed-loop control, evaluations on representative dexterous loco-manipulation motions show that our policy achieves higher manipulation performance and approximately $2\times$ faster convergence than MLP baselines. We further deploy DexWeave on a physical Unitree G1 humanoid equipped with Inspire dexterous hands, demonstrating dexterous whole-body loco-manipulation in the real world. Beyond the policies studied here, DexWeave provides a potential pathway for converting abundant human interaction data into embodiment-specific, physically executable robot data for future embodied foundation models.

\section{Related Work}

\paragraph{Learning Humanoid Skills from Human Demonstrations.}
Human demonstrations provide a scalable source of whole-body behaviors without requiring extensive robot-specific teleoperation. OmniH2O~\citep{he2024omnih2o} combines large-scale human motion retargeting with reinforcement learning to acquire whole-body humanoid skills, while VideoMimic~\citep{allshire2025videomimic} reconstructs humans and their surrounding environments from videos to learn context-aware humanoid control. More recent systems extend this direction toward whole-body manipulation. HumDex~\citep{heng2026humdex} leverages whole-body human motion to learn transferable motion priors before adapting them with robot data. These works demonstrate the potential of human data for scalable humanoid skill acquisition. DexWeave focuses on a complementary challenge: transferring demonstrations with coordinated body, hand, and object motion across embodiments while preserving the interactions needed for physically executable dexterous whole-body skills.

\paragraph{Human-to-Robot Motion Retargeting.}
Motion retargeting bridges the substantial morphology gap between human demonstrations and humanoid robots. Whole-body methods adapt human motion to robot kinematics while explicitly enforcing constraints such as joint limits, collisions, contacts, and support feasibility~\citep{araujo2025gmr,yang2025omniretarget,nvidia2026somaretargeter}. In parallel, dexterous-hand methods focus on transferring fine-grained finger articulation and hand--object relationships~\citep{handa2020dexpilot,xin2025objectives,wu2026toporetarget}. TopoRetarget~\citep{wu2026toporetarget}, for example, explicitly preserves task-relevant hand--object interaction structure, while SPIDER~\citep{pan2025spider} employs physics-informed refinement to transform kinematic human demonstrations into dynamically feasible robot trajectories. Despite this progress, whole-body and dexterous-hand retargeting are often optimized with different objectives or at separate stages. Across embodiments, differences in arm reach, palm geometry, and finger proportions can therefore make wrist placement and finger articulation mutually inconsistent even when each is locally well retargeted. This motivates preserving interaction structure across the entire upper-body interaction chain rather than only at the body or hand level. DexWeave jointly refines the arms, wrists, and fingers to maintain wrist--finger--object coordination while preserving feasible lower-body support.

\paragraph{Humanoid Loco-Manipulation.}
Humanoid loco-manipulation requires coordinating locomotion, balance, upper-body motion, and object interaction within a high-dimensional action space. Existing approaches make this problem tractable through control decomposition, unified whole-body policies, or learned motion priors. FALCON~\citep{zhang2026falcon} decomposes force-adaptive loco-manipulation into jointly trained lower-body locomotion and upper-body manipulation agents. ULC~\citep{sun2025ulc} demonstrates unified locomotion and upper-body control within a single policy, while CoorDex~\citep{li2026coordex} extends loco-manipulation to dexterous hands using separately trained and distilled body and hand priors coordinated through downstream residual reinforcement learning.
More recently, vision--language--action (VLA) and world-action models (WAMs) have extended humanoid loco-manipulation toward generalizable visuomotor control. WholeBodyVLA~\citep{jiang2026wholebodyvla} leverages action-free human videos to alleviate the scarcity of humanoid loco-manipulation data, while OpenHLM~\citep{hu2026openhlm} maps visual and language observations to whole-body humanoid actions. MotionWAM~\citep{zheng2026motionwam}, $\omega$-0~\citep{li2026omega0}, and related work~\citep{li2026wholebodywam} further exploit large-scale visual or motion priors for whole-body action generation. These developments expose a complementary data bottleneck: human motion and interaction data are abundant, yet embodiment-specific, physically executable whole-body dexterous trajectories remain costly to acquire. DexWeave addresses this conversion gap by grounding coordinated human body--hand--object demonstrations into interaction-consistent, physically executable robot behaviors, providing a complementary source of whole-body supervision for future embodied models.

\section{DexWeave}

\subsection{Overview}\label{sec:method_overview}

\begin{figure}[t]
\centering
\includegraphics[width=\textwidth]{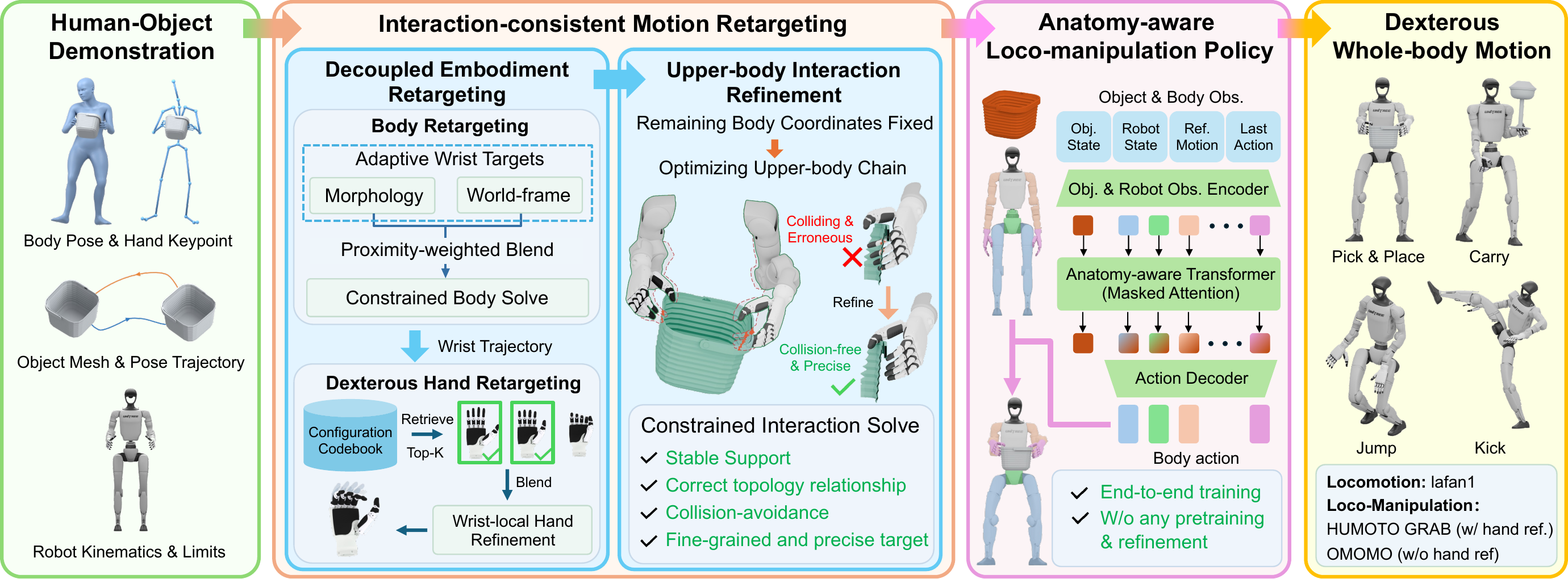}
\caption{\textbf{Overview of DexWeave.} Given a human body--hand--object demonstration, DexWeave first constructs an interaction-consistent robot reference through specialized body--hand initialization and coupled refinement of the upper-body interaction chain. An anatomy-aware whole-body loco-manipulation policy then realizes the reference under robot dynamics using regional anatomical tokens, directed masked attention, and selective object conditioning, producing executable dexterous humanoid loco-manipulation skills.}
\label{fig:overview}
\end{figure}

Given a human body--hand--object demonstration, DexWeave converts it into a physically executable dexterous humanoid skill by preserving interaction structure across embodiments and exploiting anatomical structure for control. As shown in Fig.~\ref{fig:overview}, the framework first constructs an interaction-consistent robot reference and then learns a policy to execute it under physical robot dynamics. \textbf{Interaction-consistent motion retargeting} uses specialized body--hand initialization and coupled upper-body refinement to preserve hand--object relationships while maintaining the established support motion (Sec.~\ref{sec:retargeting}). \textbf{Anatomy-aware loco-manipulation policy learning} then uses regional tokens, directed information flow, and selective object conditioning to coordinate locomotion and dexterous manipulation (Sec.~\ref{sec:whole_body_control}).

\subsection{Interaction-Consistent Motion Retargeting}\label{sec:retargeting}
Retargeting human demonstrations to a robot must reconcile whole-body reachability, hand\textendash{}object contact fidelity, and physical feasibility. Optimizing all three jointly is ill-conditioned, because body-scale differences, finger-dimension mismatches, and collision constraints interact across the full kinematic chain. DexWeave therefore decomposes the problem according to interaction locality. A decoupled initialization stage first establishes body motion and hand shape where each is locally well-posed (Sec. \ref{sec:decoupled_retargeting}), and a subsequent refinement stage closes the residual misalignment on the compact upper-limb chain that mediates contact (Sec. \ref{sec:upper_limb_refinement}).


A demonstration provides human body poses $\mathbf X_t^H$, hand keypoints $\mathbf H_t^s$ for $s\in\{L,R\}$, and rigid-object poses $\mathbf X_t^O$ over frames $t=1,\ldots,T$. The robot configuration $\mathbf q_t=(\mathbf q_t^B,\boldsymbol\theta_t^L,\boldsymbol\theta_t^R)$ consists of the floating base and non-finger joints $\mathbf q_t^B$ together with the independent hand drivers $\boldsymbol\theta_t^s$, from which dependent finger joints are determined by the robot's mimic relations. The pipeline first constructs a decoupled initialization $\bar{\mathbf q}_{1:T}$ and then refines its upper-body coordinates to produce $\mathbf q_{1:T}^*$. Together with the original object trajectory, this yields the reference $\mathcal R_{1:T}={\left\{(\mathbf q_t^*,\mathbf X_t^O)\right\}}_{t=1}^{T}$ for downstream policy learning.

\subsubsection{Decoupled Body\textendash{}Hand Initialization}\label{sec:decoupled_retargeting}

The initialization stage solves body motion and hand articulation using specialized solvers before coupling them during interaction refinement. This decomposition reflects the different structure of the two subproblems: body retargeting is dominated by large-scale morphology adaptation and support constraints, whereas hand fitting depends on local finger geometry and kinematic coupling. Solving each component in the representation where it is locally well posed provides a robust initialization for subsequent joint refinement. 

\paragraph{Contact-Aware Body Retargeting.}
The wrist is the point at which whole-body retargeting most directly affects manipulation. Rescaling human motion to robot proportions can displace the wrist relative to nearby objects and break contact even when the overall body pose is well retargeted. We therefore blend two wrist targets: a robot-scaled morphology-preserving target~\citep{araujo2025gmr} and a world-aligned interaction-preserving target. Their contribution is controlled by a per-hand weight $\alpha_{t,s}\in[0,1]$ determined by persistent hand--object proximity. When the hand is far from the object, the target favors morphology adaptation; as sustained interaction emerges, the blend gradually shifts toward scene alignment. This provides a smooth transition without introducing a discrete contact mode switch. Wrist orientations remain fully mapped and the object trajectory is kept unchanged. Support constraints inferred from source foot kinematics suppress unintended foot sliding, while clearance penalties handle self-, ground-, and body--object collisions. At each frame, the body configuration is obtained by solving:
\begin{equation}
\bar{\mathbf q}_t^B
= \arg\min_{\mathbf q^B \in \mathcal Q_t^B}
\lambda_{\mathrm{track}}\mathcal L_{\mathrm{track}}
+\lambda_{\mathrm{clr}}\mathcal L_{\mathrm{clr}}
+\lambda_{\mathrm{temp}}\mathcal L_{\mathrm{temp}}
+\mathcal L_{\mathrm{reg}},
\label{eq:whole_body_retargeting}
\end{equation}
where $\mathcal L_{\mathrm{track}}$ penalizes deviation from body landmarks and the blended wrist targets $\widehat{\mathcal W}_t$, $\mathcal L_{\mathrm{clr}}$ collects soft collision terms, $\mathcal L_{\mathrm{temp}}$ regularizes temporal posture, and $\mathcal L_{\mathrm{reg}}$ denotes a regularization term. The feasible set $\mathcal Q_t^B$ encodes joint limits, support equalities, and body-collision boundaries.

\paragraph{Retrieval-Based Dexterous Hand Fitting.}
Nonlinear finger kinematics and mimic coupling make direct geometric fitting sensitive to initialization~\citep{handa2020dexpilot,xin2025objectives}. We address this using a precomputed codebook
$\mathcal B^s={(\boldsymbol\theta_n^s,\mathbf g_n^{s,R})}_{n=1}^{N_B}$,
which pairs feasible hand-driver configurations with wrist-local geometric descriptors. These descriptors encode fingertip positions, chain directions, and inter-finger relations computed through forward kinematics.
Given observed human hand keypoints, we construct the corresponding descriptor $\mathbf g_t^{s,H}$ in the same wrist-local frame and retrieve the $K$ nearest codebook entries. Their driver configurations are blended using normalized distance weights to obtain $\widehat{\boldsymbol\theta}_t^s$, which serves as both a warm start and a soft anchor for continuous fitting of the independent drivers within their mimic-consistent bounds. Sequences without articulated hand references default to a neutral configuration $\boldsymbol\theta_{\mathrm{neutral}}^s$. Combining the fitted hands with the retargeted body trajectory yields the whole-body initialization $\bar{\mathbf q}_{1:T}$.


\subsubsection{Coupled Upper-Body Interaction Refinement}\label{sec:upper_limb_refinement}

The previous initialization can still exhibit residual hand--object misalignment because embodiment differences propagate jointly through arm reach, palm geometry, and finger proportions. Correcting such errors only at the wrist or fingers can therefore distort another part of the interaction chain. DexWeave instead jointly refines the upper-body kinematic chain that governs hand--object interaction. In particular, we define a compact set of joint variables $\mathbf z_t=[\mathbf q_t^U,\boldsymbol\theta_t^L,\boldsymbol\theta_t^R]^\top$ that contains the bilateral arm and wrist joints (i.e. $\mathbf q_t^U$) together with both hands' independent drivers (i.e. $\boldsymbol\theta_t^L,\boldsymbol\theta_t^R$). All other body joints retain their initialized values in $\bar{\mathbf q}_t$, preserving the lower-body support motion.

Three design choices shape this refinement. \textit{(i)~Compact active set.} Restricting optimization to the upper-body interaction chain confines corrections to joints that directly influence contact and prevents unnecessary drift in lower-body support. \textit{(ii)~Priority-weighted fingertip alignment.} Following precision-grasping principles~\citep{handa2020dexpilot}, we assign larger weights to the thumb and index fingertips, which often dominate interaction geometry, while the remaining fingertips provide complementary grasp and support cues. \textit{(iii)~Relaxed wrist anchoring.} Rather than fixing the wrist at its initialized position, we soften its positional anchor while retaining palm-orientation objectives~\citep{xin2025objectives}. This allows residual mismatch to be distributed across wrist placement and finger articulation instead of forcing the fingers alone to absorb the correction.

Let $\bar{\mathbf z}_t$ denote the initialized active coordinates. At each frame, the refinement solves
\begin{equation}
\mathbf z_t^*
= \arg\min_{\mathbf z \in \mathcal Z_t}
\lambda_{\mathrm{tip}}\mathcal L_{\mathrm{tip}}
+\lambda_{\mathrm{ori}}\mathcal L_{\mathrm{ori}}
+\lambda_{\mathrm{clr}}\mathcal L_{\mathrm{clr}}
+\lambda_{\mathrm{temp}}\mathcal L_{\mathrm{temp}}
+\mathcal L_{\mathrm{reg}},
\label{eq:upper_limb_objective}
\end{equation}
where $\mathcal L_{\mathrm{tip}}$ penalizes priority-weighted fingertip position error, $\mathcal L_{\mathrm{ori}}$ tracks the palm normal, wrist-forward direction, and hand-lateral direction, $\mathcal L_{\mathrm{clr}}$ collects soft collision-clearance and recovery penalties, $\mathcal L_{\mathrm{temp}}$ regularizes the correction $\mathbf z_t-\bar{\mathbf z}_t$ over time, , and $\mathcal L_{\mathrm{reg}}$ denotes a regularization term. Replacing the active joint values of $\bar{\mathbf q}_t$ with $\mathbf z_t^*$ yields the refined whole-body trajectory $\mathbf q_{1:T}^*$. Through this coupled refinement, wrist placement and finger articulation are optimized as a single interaction structure rather than as independent targets. Loss decompositions, dataset-specific target assignments, and the complete definitions of $\mathcal Q_t^B$ and $\mathcal Z_t$ are provided in Appendix~\ref{app:retargeting_protocol}.


\subsection{Anatomy-Aware Loco-Manipulation Policy Learning}
\label{sec:whole_body_control}

\begin{figure}[t]
\centering
\includegraphics[width=\textwidth]{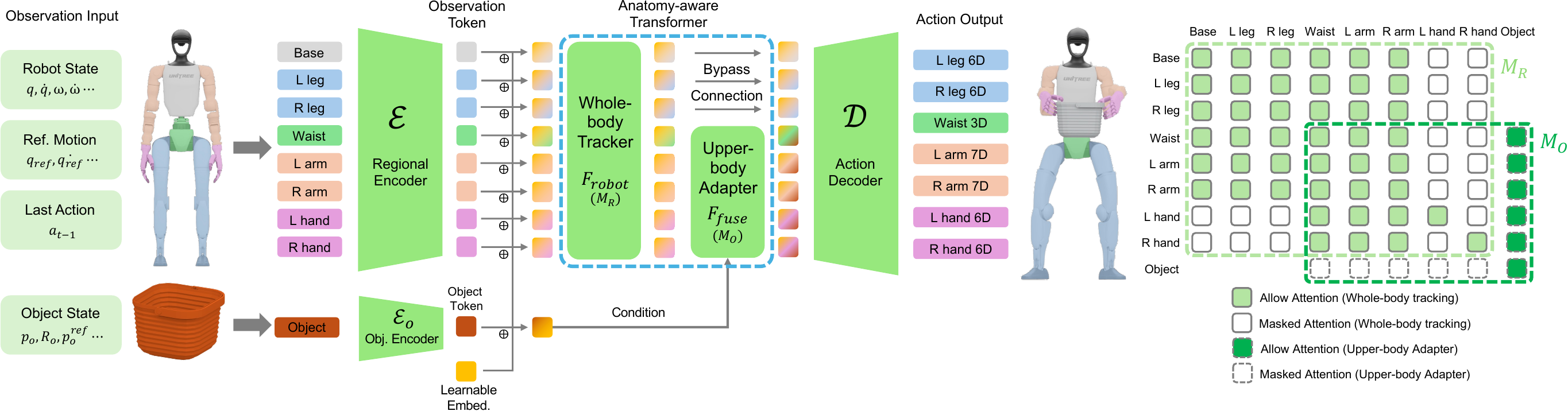}
\caption{\textbf{Anatomy-aware policy architecture.} \textbf{Left:} Robot observations are factorized into anatomical tokens and
coordinated by a masked whole-body Transformer. Object information is fused
one-way into the waist, arm, and hand branches, while the original waist feature
is retained for object-independent leg control. Seven group-specific action
heads produce the full body-and-hand action.
\textbf{Right:} Attention-mask visualization. Rows attend to columns. Light and
dark green indicate allowed attention in the robot Transformer and object-fusion
Transformer, respectively.}\label{fig:rl_policy}
\end{figure}

Retargeting provides interaction-consistent references, but realizing them under
robot dynamics requires coordinated control of locomotion, whole-body posture,
and articulated hands. DexWeave explicitly incorporates humanoid anatomy into
both representation and information flow using an asymmetric actor--critic
architecture. The actor receives a 238-dimensional observation
$\mathbf o_t=[\mathbf o_t^{\mathrm{robot}},
\mathbf o_t^{\mathrm{object}}]$, comprising a 211-dimensional robot observation
and a 27-dimensional object observation, while the critic uses privileged
simulation observations during training. The network architecture is shown in Figure ~\ref{fig:rl_policy} .

\subsubsection{Regional Tokenization and Directed Coordination}
\label{sec:body_tokenization}

The robot observation is factorized into eight anatomical regions corresponding
to the base, bilateral legs, waist, bilateral arms, and bilateral hands.
Region-specific encoders $\mathcal E_i$ map each regional observation
$\mathbf x_i$ into a shared $d$-dimensional space ($d=128$), with a learned
body-part embedding $\mathbf p_i$:
\begin{equation}
\mathbf h_i=\mathcal E_i(\mathbf x_i)+\mathbf p_i,\qquad
\mathbf H=[\mathbf h_1,\ldots,\mathbf h_8]^\top .
\label{eq:anatomy_token}
\end{equation}

A two-layer masked Transformer produces
\begin{equation}
\mathbf Z=F_{\mathrm{robot}}(\mathbf H;M_R),
\label{eq:robot_transformer}
\end{equation}
where the anatomical mask $M_R$ allows bidirectional communication among the
base, legs, waist, and arms, while preventing body tokens from attending to the
hands. Each hand token attends to the waist, both arms, and itself. This
directed structure provides finger control with upstream body context without
feeding hand-specific features back into the body-control representation.

\subsubsection{Selective Object Conditioning}
\label{sec:directed_attention}

The object observation contains the current pose, reference pose, and their
discrepancy in the torso frame, where each pose uses a 3D position and a 6D
rotation representation from the first two columns of the rotation matrix.
It is encoded as
\begin{equation}
\mathbf h_o
=
\mathcal E_o(\mathbf o_t^{\mathrm{object}})+\mathbf p_o ,
\end{equation}
where $\mathbf p_o$ is a learned object-token embedding.

Object information is fused only into the waist, arm, and hand tokens.
Let $\mathcal I_U$ denote these five regions. A one-layer fusion Transformer
computes
\begin{equation}
\left[
\widetilde{\mathbf Z}^{\mathcal I_U},
\widetilde{\mathbf h}_o
\right]
=
F_{\mathrm{fuse}}
\left(
[\mathbf Z^{\mathcal I_U},\mathbf h_o];M_o
\right),
\label{eq:object_conditioned_actor}
\end{equation}
where each selected robot token attends only to itself and the object token,
and the object token attends only to itself. The base and leg tokens bypass
object fusion. Importantly, the original object-independent waist token is
retained for the leg-action heads, making leg actions structurally independent
of direct object observations.

\subsubsection{Grouped Action Decoding}
\label{sec:action_decoding}

Seven group-specific MLP heads decode the structured features into a
41-dimensional action mean. Each head receives anatomically relevant context:
leg heads use the base, corresponding leg, and object-independent waist tokens;
the waist and arm heads use object-conditioned upper-body features; and each
hand head uses the corresponding arm and hand features. Collectively,
\begin{equation}
\boldsymbol\mu_t
=
\mathcal D
\left(
\mathbf Z_t,
\widetilde{\mathbf Z}_t^{\mathcal I_U}
\right).
\label{eq:grouped_decoder}
\end{equation}
The actor parameterizes a diagonal Gaussian policy,
\begin{equation}
\pi_\theta(\mathbf a_t\mid\mathbf o_t)
=
\mathcal N
\left(
\boldsymbol\mu_t,
\operatorname{diag}(\boldsymbol\sigma^2)
\right),
\end{equation}
with independently learned standard deviations. The 41 outputs represent normalized joint-position commands rather than joint torques. They are converted to joint-position targets using joint-specific scales and offsets and subsequently executed by joint-level PD controllers; the 12 independent hand commands are expanded through fixed mimic-joint relations.

\subsubsection{Single-Stage Actor--Critic Learning}
\label{sec:ppo_training}

All actor components are jointly optimized with PPO~\citep{schulman2017ppo}.
Tracking rewards supervise the retargeted body and hand motion, while
manipulation rewards supervise object interaction. The actor uses deployable
observations, including noisy and delayed object measurements, whereas the
critic receives privileged simulation state. DexWeave therefore learns
whole-body tracking and dexterous interaction in a single training stage,
without tracking-policy pretraining, teacher--student distillation, or
subsequent residual-policy refinement. Further implementation and training
details are provided in Appendix~\ref{app:policy_protocol}.

\section{Experiments}\label{sec:experiments}

We evaluate DexWeave at two levels on a Unitree G1 with Inspire hands: retargeting quality and closed-loop control. Section~\ref{sec:retargeting_experiments} evaluates whole-body feasibility and hand--object geometry, while Section~\ref{sec:policy_experiments} examines whether the retargeted references enable effective policy learning. Ablation results are reported in Appendix~\ref{sec:ablation}.


\subsection{Retargeting Evaluation}\label{sec:retargeting_experiments}

We evaluate DexWeave across progressively richer interaction settings: body motion in LAFAN1~\citep{harvey2020robust}, whole-body object interaction without finger references in OMOMO~\citep{li2023object}, and articulated hand--object interaction in GRAB~\citep{taheri2020grab} and HUMOTO~\citep{lu2025humoto}. This progression evaluates whether whole-body support remains feasible as increasingly detailed interaction geometry is introduced. Appendix~\ref{app:retargeting_protocol} provides the details on retargeting settings, and Appendix~\ref{app:evaluation} defines the evaluation metrics and aggregation protocol.

\paragraph{Whole-body feasibility and coarse contact preservation.}

LAFAN1 and OMOMO evaluate whole-body feasibility before introducing fine-grained finger geometry, with OMOMO additionally providing hand--object contact information. Because OMOMO does not contain observed fingertip targets, hand refinement primarily encourages contact between the robot hand and object surfaces. We compare DexWeave with OmniRetarget~\citep{yang2025omniretarget}, GMR~\citep{araujo2025gmr}, and SOMA~\citep{nvidia2026somaretargeter}. Table~\ref{tab:retarget_body_contact} shows the lowest penetration and near-zero foot skating for DexWeave on both datasets. On OMOMO, contact duration reaches 0.999 and contact distance falls to 2.944\,cm, compared with 0.879 and 7.677\,cm for the strongest contact baseline, GMR. The contact gain is accompanied by a penetration duration of only 0.002, showing that closer hand\textendash{}object alignment is achieved while retaining whole-body feasibility under the measured criteria.

\begin{table}[!htb]
    \centering
    \caption{
        \textbf{Whole-body feasibility and coarse interaction preservation on LAFAN1 and OMOMO.}
        Penetration and foot skating measure feasibility; contact preservation is evaluated only on OMOMO. Durations are fractions, and $\sim\!0$ denotes a near-zero value. \bestvalue{Yellow} and \underline{underlined} mark the best and second-best values within each dataset.
    }
    \label{tab:retarget_body_contact}

    \footnotesize
    \setlength{\tabcolsep}{4pt}
    \begin{tabular*}{\linewidth}{@{\extracolsep{\fill}}llcccccc}
        \toprule
        & & \multicolumn{2}{c}{\textbf{Penetration}}
        & \multicolumn{2}{c}{\textbf{Foot Skating}}
        & \multicolumn{2}{c}{\textbf{Contact Preservation}} \\

        \cmidrule(lr){3-4}
        \cmidrule(lr){5-6}
        \cmidrule(lr){7-8}

        \textbf{Dataset} & \textbf{Method}
        & \shortstack{Duration \\ $\downarrow$}
        & \shortstack{Max Depth \\ (cm) $\downarrow$}
        & \shortstack{Duration \\ $\downarrow$}
        & \shortstack{Max Vel. \\ (m/s) $\downarrow$}
        & \shortstack{Duration \\ $\uparrow$}
        & \shortstack{Distance \\ (cm) $\downarrow$} \\

        \midrule
        \multirow{4}{*}{\textit{LAFAN1}} & OmniRetarget
        & $\underline{0.125}$ & $\underline{2.934}$
        & 0.091 & $\underline{0.416}$ & N/A & N/A \\

        & GMR
        & 0.183 & 4.314
        & $\underline{0.010}$ & 0.548 & N/A & N/A \\

        & SOMA
        & 0.968 & 6.379
        & 0.115 & 0.565 & N/A & N/A \\

        & DexWeave (Ours)
        & \bestvalue{$\sim 0$} & \bestvalue{$2.575$}
        & \bestvalue{$0$} & \bestvalue{$0$} & N/A & N/A \\

        \midrule
        \multirow{4}{*}{\textit{OMOMO}} & OmniRetarget
        & $\underline{0.131}$ & $\underline{2.735}$
        & $\underline{\sim{}0}$ & $1.455$ & 0.644 & 10.289 \\

        & GMR
        & 0.916 & 3.117
        & 0.008 & 0.802 & $\underline{0.879}$ & $\underline{7.677}$ \\

        & SOMA
        & 0.746 & 4.163
        & 0.034 & $\underline{0.467}$ & 0.577 & 16.205 \\

        & DexWeave (Ours)
        & \bestvalue{$0.002$} & \bestvalue{$2.117$}
        & \bestvalue{$\sim{0}$} & \bestvalue{$0.355$}
        & \bestvalue{$0.999$} & \bestvalue{$2.944$} \\

        \bottomrule
    \end{tabular*}
\end{table}


\begin{table}[!htb]
    \centering
    \caption{
        \textbf{Fine-grained interaction retargeting on GRAB and HUMOTO.} Penetration duration is the fraction of colliding frames. Primary and secondary errors measure thumb/index and remaining-finger positions; palm error measures normal alignment. Lower is better. \bestvalue{Yellow} and \underline{underlined} mark the best and second-best values within each dataset. Full comparison can be found in Appendix~\ref{app:retargeting_results}. 
    }
    \label{tab:retarget_dexterous}

    \small
    \setlength{\tabcolsep}{4pt}

    \begin{tabular*}{\linewidth}{@{\extracolsep{\fill}}llccccc}
        \toprule
        & & \multicolumn{2}{c}{\textbf{Penetration}}
        & \multicolumn{3}{c}{\textbf{Hand Alignment}} \\

        \cmidrule(lr){3-4}
        \cmidrule(lr){5-7}

        \textbf{Dataset} & \textbf{Method}
        & \shortstack{Duration \\ $\downarrow$}
        & \shortstack{Max Depth \\ (cm) $\downarrow$}
        & \shortstack{Pri. Err. \\ (mm) $\downarrow$}
        & \shortstack{Sec. Err. \\ (mm) $\downarrow$}
        & \shortstack{Palm Err. \\ (°) $\downarrow$} \\

        \midrule
        \multirow{3}{*}{\textit{GRAB}} & OmniRetarget + DexPilot \textbf{+ IK}
        & 0.212 & $\underline{2.269}$ & 16.585 & \bestvalue{$13.368$} & 11.555 \\

        & OmniRetarget + SBR \textbf{+ IK}
        & $\underline{0.190}$ & 2.362 & $\underline{14.925}$ & 14.659 & $\underline{9.530}$ \\

        & DexWeave (Ours)
        & \bestvalue{$0.001$} & \bestvalue{$1.263$} & \bestvalue{$4.342$} & $\underline{14.476}$ & \bestvalue{$3.652$} \\

        \midrule
        \multirow{3}{*}{\textit{HUMOTO}} & OmniRetarget + DexPilot \textbf{+ IK}
        & 0.522 & 3.597 & 32.389 & 31.018 & 14.444 \\

        & OmniRetarget + SBR \textbf{+ IK}
        & $\underline{0.511}$ & $\underline{3.582}$ & $\underline{29.999}$ & $\underline{28.436}$ & $\underline{12.476}$ \\

        & DexWeave (Ours)
        & \bestvalue{$0.007$} & \bestvalue{$2.854$} & \bestvalue{$7.198$} & \bestvalue{$11.702$} & \bestvalue{$4.439$} \\

        \bottomrule
    \end{tabular*}
\end{table}

\begin{figure}[t]
    \centering
    \includegraphics[width=\linewidth]{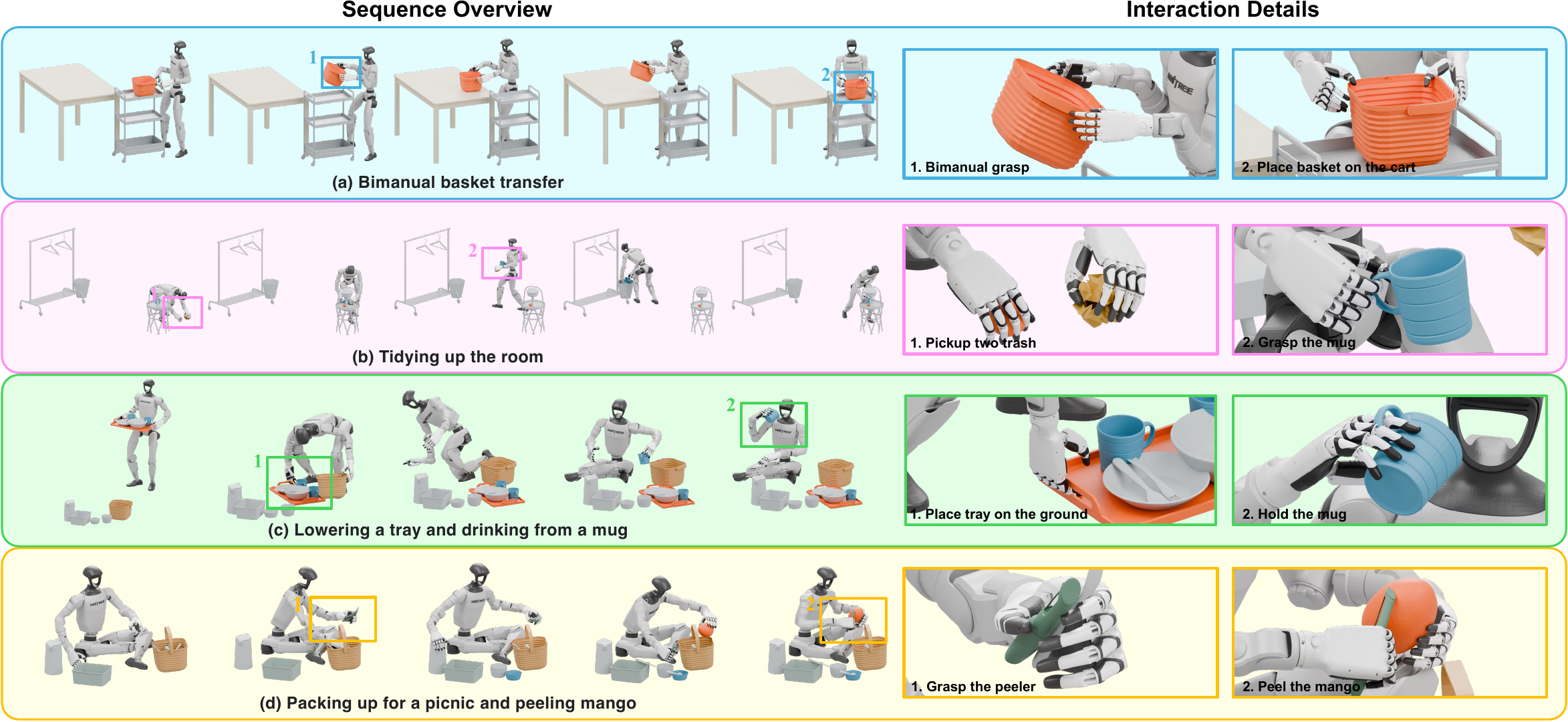}
    \caption{
        \textbf{Qualitative retargeting results across interaction levels.}
        Each row shows a sequence overview (left) and hand\textendash{}object close-ups (right).
    }\label{fig:retargeting_results_main}
    \vspace{-3mm}
\end{figure}


\paragraph{Fine-grained interaction retargeting.}

We compare DexWeave with OmniRetarget paired with DexPilot~\citep{handa2020dexpilot} or SBR~\citep{malate2026smoothoperator}, each followed by arm\textendash{}wrist IK that adjusts hand placement while keeping finger articulation fixed. Table~\ref{tab:retarget_dexterous} shows that DexWeave lowers primary-fingertip error to 4.342\,mm on GRAB and 7.198\,mm on HUMOTO, compared with 14.925 and 29.999\,mm for the best IK variants, while achieving the lowest penetration duration and palm error. Its secondary-fingertip error ranks first on HUMOTO and second on GRAB. These joint gains are consistent with refining the arm, wrist, and fingers as one interaction chain. Figure~\ref{fig:retargeting_results_main} shows representative whole-body motions and hand\textendash{}object close-ups.

\subsection{Policy Evaluation}\label{sec:policy_experiments}

The policy study asks whether the retargeted references can be executed under dynamics, first in body-only tracking and then in dexterous loco-manipulation. We train one policy per reference and compare actors with matched task-specific observations, actions, rewards, reset and termination rules, domain randomization, and training budgets. Reported values summarize the common final training window for each comparison;
Appendices~\ref{app:policy_protocol} and~\ref{app:evaluation} provide the details on training and aggregation settings.

\paragraph{Motion tracking.}

\begin{figure}[!htb]
\centering
\begin{minipage}[t]{0.50\linewidth}
    \centering
    \vspace{0pt}
    \begin{minipage}[t][5.15cm][t]{\linewidth}
    \centering
    \scriptsize
    \setlength{\tabcolsep}{2pt}
    \resizebox{\linewidth}{!}{%
    \begin{tabular}{@{}lcccc@{}}
        \toprule
        \multicolumn{5}{c}{\textbf{Body-only motion tracking}} \\
        \cmidrule(lr){1-5}
        \textbf{Actor}
        & \shortstack{Complete Ratio\\(\%) $\uparrow$}
        & \shortstack{Body pos.\\(cm) $\downarrow$}
        & \shortstack{Anchor pos.\\(cm) $\downarrow$}
        & \shortstack{Anchor rot.\\(deg) $\downarrow$}\\

        \midrule

        Any2Track
        & 72.82 & 6.09 & 44.88 & 32.07  \\

        GMT
        & 69.86 & 5.85 & 36.06 & 21.92  \\

        BeyondMimic
        & \textbf{100} & 2.82 & 5.19 & 2.48  \\

        DexWeave (Ours)
        & \textbf{100} & \textbf{2.62} & \textbf{4.82} & \textbf{2.34}  \\

        \midrule
        \multicolumn{5}{c}{\textbf{Dexterous loco-manipulation}} \\
        \cmidrule(lr){1-5}
        \textbf{Actor}
        & \shortstack{Success rate\\(\%) $\uparrow$}
        & \shortstack{Body pos.\\(cm) $\downarrow$}
        & \shortstack{Object pos.\\(cm) $\downarrow$}
        & \shortstack{Object rot.\\(deg) $\downarrow$} \\

        \midrule
        InterMimic
        & 57.74 & 6.51 & 3.58 & 7.71  \\

        Object MLP
        & 85.00 & 5.69  & 3.84 & \textbf{6.16} \\

        DexWeave (Ours)
        & \textbf{97.50} & \textbf{4.93}  & \textbf{3.43} & 6.83  \\

        \bottomrule
    \end{tabular}%
    }
    \end{minipage}
    \vspace{-3mm}
    \captionof{table}{\textbf{Policy performance under MuJoCo sim-to-sim evaluation.} Results are averaged equally over five tracking motions and five loco-manipulation references. Any2Track and GMT serve as general motion tracking baselines. Following InterMimic's official human-model setup, we evaluate it using an SMPL-X skeleton as a cross-embodiment reference. DexWeave and Object MLP share the G1--Inspire embodiment. Completion and success rates indicate the percentage of evaluation episodes reaching the reference end or time limit. Best values are in \textbf{bold}.}\label{tab:policy_results}
\end{minipage}\hfill
\begin{minipage}[t]{0.48\linewidth}
    \centering
    \vspace{0pt}
    \begin{minipage}[t][5.15cm][t]{\linewidth}
    \centering
    \includegraphics[width=\linewidth]{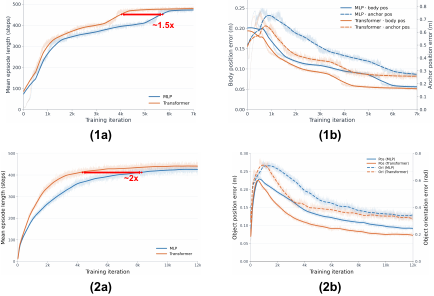}
    \end{minipage}
    \vspace{-3mm}
    \captionof{figure}{\textbf{Policy learning curves.} The top row shows tracking episode length (1a) and body-position (solid) and anchor-position (dashed) errors (1b); the bottom row shows loco-manipulation episode length (2a) and object-position (solid) and orientation (dashed) errors (2b). Blue denotes the MLP and orange denotes DexWeave.}\label{fig:policy_learning}
\end{minipage}
\end{figure}

To isolate body coordination without object conditioning, we compare DexWeave with Any2Track ~\citep{zhang2026any2track}, GMT ~\citep{chen2025gmt}, and BeyondMimic~\citep{liao2025beyondmimic} on LAFAN1 motions~\citep{harvey2020robust}. As shown in Table~\ref{tab:policy_results}, DexWeave achieves a 100\% completion ratio, matching BeyondMimic while outperforming all baselines in tracking accuracy. Compared with BeyondMimic, DexWeave reduces body-position error from 2.82 to 2.62 \,cm, torso-anchor position error from 5.19 to 4.82 \,cm, and anchor rotation error from 2.48$^\circ$ to 2.34$^\circ$. Any2Track and GMT achieve lower completion ratios of 72.82\% and 69.86\%, respectively. 

\paragraph{Dexterous loco-manipulation.}
The main control evaluation combines whole-body motion with articulated hand--object interaction on reference motions. We compare DexWeave with InterMimic \citep{xu2025intermimic} and an object-conditioned MLP baseline, i.e. we replace DexWeave's Transformer network with an MLP of comparable parameter count. As shown in Table~\ref{tab:policy_results}, DexWeave achieves the highest success rate of 97.5\%, compared with 85.0\% for the MLP baseline and 57.74\% for InterMimic. It also attains the lowest body-position error (4.93\,cm) and object-position error (3.43\,cm). The MLP baseline achieves a slightly lower object-rotation error (6.16$^\circ$ versus 6.83$^\circ$). Overall, DexWeave improves task success while maintaining more accurate body and object tracking. Figure~\ref{fig:policy_learning} shows the corresponding training curves, further demonstrating that DexWeave converges faster than its MLP counterpart.

\begin{figure}[!htb]
    \centering
    \includegraphics[width=\linewidth]{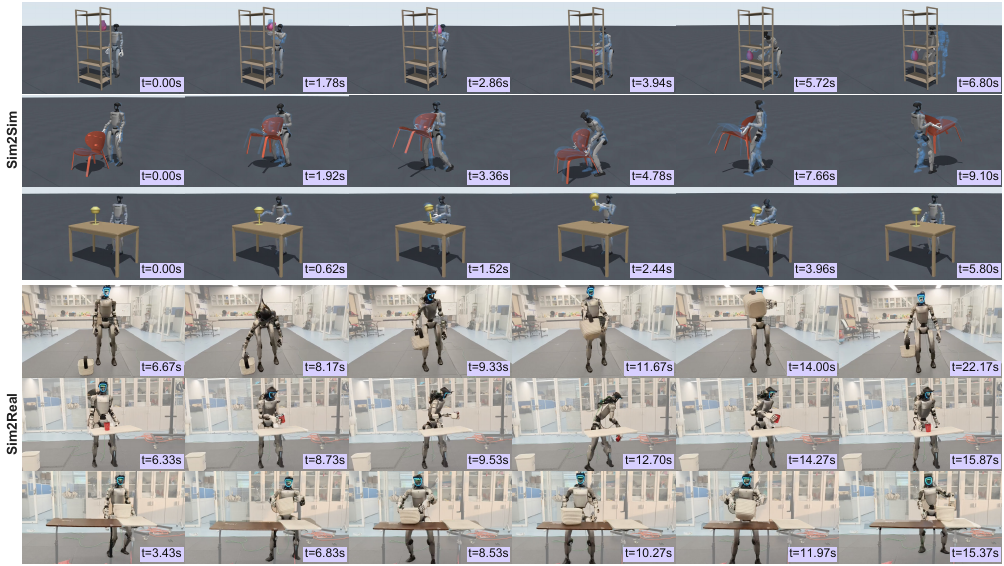}
    \caption{
        \textbf{Sim-to-sim and sim-to-real transfer of dexterous loco-manipulation policies.}
        Policies are trained in IsaacLab, evaluated in MuJoCo for sim-to-sim transfer, and deployed on a physical Unitree G1 with Inspire dexterous hands for sim-to-real transfer.
    }\label{fig:execution_results}
\end{figure}

\paragraph{Sim-to-sim and hardware evaluation.}
We evaluate DexWeave through both sim-to-sim and sim-to-real transfer on several dexterous loco-manipulation tasks. Figure~\ref{fig:execution_results} shows corresponding executions across simulation environments and on a physical Unitree G1 equipped with Inspire dexterous hands. The results show that the learned policies preserve coordinated whole-body motion and dexterous object interaction across simulators and transfer to the physical platform, providing qualitative evidence that the human-derived references can be realized as executable robot skills.

\section{Conclusion}


We introduced \textbf{DexWeave}, which transfers human demonstrations into executable dexterous humanoid loco-manipulation skills by combining interaction-consistent retargeting with anatomy-aware policy learning. Experiments in simulation and on a physical humanoid demonstrate its effectiveness in generating embodiment-specific training data for future embodied foundation models.

\bibliography{iclr2027_conference_updated}
\bibliographystyle{iclr2027_conference}

\clearpage
\appendix

\section{Retargeting Details}\label{app:experimental_details}\label{app:retargeting_protocol}

\subsection{Reference Preprocessing and Dataset Targets}

\paragraph{Time and coordinate conventions.}
All human, robot, and object trajectories are represented at 50\,Hz ($\Delta t=0.02$\,s), preserving their original timing. Using zero-based frame indices, output frame $k$ corresponds to the fractional source index $k f_{\mathrm{src}}/50$, where $f_{\mathrm{src}}$ is the source frame rate. Rotations are interpolated on $\mathrm{SO}(3)$. A shared heading alignment and ground-height translation are applied to all trajectories and world-space targets. The original scene scale is preserved by default ($s_{\mathrm{scene}}=1$). When real-world interaction targets exceed the robot's reach, we adjust the scene scale only as needed, applying $\mathbf p'=\mathbf a+s_{\mathrm{scene}}(\mathbf p-\mathbf a)$ to the human and all objects about a common ground anchor $\mathbf a$. Object mesh dimensions use the same scale factor; their orientations are unaffected by scaling. The resulting object trajectories remain fixed throughout retargeting.

\paragraph{Interaction-dependent targets.}
For sequences with articulated hand observations, let $\mathbf p^{\mathrm{morph}}_{t,s}$ denote the morphology-adapted wrist target and $\mathbf p^{\mathrm{scene}}_{t,s}$ the corresponding wrist position in the preprocessed scene, for hand $s\in\{L,R\}$. The wrist position target is
\begin{equation}
    \widehat{\mathbf p}_{t,s}
    =
    (1-\alpha_{t,s})\mathbf p^{\mathrm{morph}}_{t,s}
    +
    \alpha_{t,s}\mathbf p^{\mathrm{scene}}_{t,s}.
\end{equation}
The morphology-adapted target follows the torso. The scene target follows the preprocessed demonstration in the shared world frame and is not modified by retargeting. The blend applies only to wrist positions,while wrist orientation targets retain their mapped values. The interaction weight $\alpha_{t,s}\in[0,1]$ is determined from persistent hand--object proximity, using a 5\,cm threshold and a nominal persistence/transition interval of $4/30$\,s, discretized to seven frames at 50\,Hz. GRAB and HUMOTO additionally use source finger-contact labels to activate scene tracking, with a smooth proximity-based ramp during approach. The same interaction weight balances world-frame fingertip alignment and wrist-local hand-shape preservation during refinement.

\paragraph{Dataset-specific targets.}
GRAB and HUMOTO provide articulated hand observations for fingertip and hand-orientation fitting. Their evaluation masks are defined directly from source finger-contact labels, independently of the smoothed target activations used during optimization. OMOMO provides body--object motion without observed fingertip targets, while its SMPL-H joint centers are therefore not treated as fingertip observations. Instead, refinement retains the observed wrist positions and initialized hand orientations, and uses robot-surface contact targets in place of fingertip tracking. LAFAN1 provides body-only motion, which fingers remain in a neutral configuration, and refinement updates only the arm and wrist joints. The corresponding refinement objectives and contact-construction details are provided in Appendix~\ref{sec:appendix_coupled}.

\subsection{Decoupled Body--Hand Initialization Settings}
\label{app:decoupled_initialization}

\subsubsection{Body Optimization and Support Constraints}
\label{app:body_support}

The body solver in Eq.~\ref{eq:whole_body_retargeting} optimizes the floating base and non-finger joints, with finger joints fixed. Table~\ref{tab:body_objectives} summarizes its soft objectives. The feasible set is
\begin{equation}
\mathcal Q_t^B =
\left\{
\mathbf q^B:
\begin{aligned}
&\mathbf q^-\leq\mathbf q_{\mathrm{act}}^B\leq\mathbf q^+,\\
&b_s(\mathbf q^B)\geq0
    \quad(s\in\{L,R\}),\\
&\mathbf c_{s,xy}(\mathbf q^B)=\mathbf a_{t,s}
    \quad(s\in\mathcal S_t),\\
&d_c(\mathbf q^B)\geq-\epsilon_c
    \quad(c\in\mathcal C_t^B)
\end{aligned}
\right\},
\label{eq:app_body_constraints}
\end{equation}
where $\mathbf q_{\mathrm{act}}^B$ contains the actuated body coordinates. The elbow constraint preserves the physical bending side through
\begin{equation*}
b_s =
\frac{
\mathbf n_s^\top
\left[
(\mathbf p_w-\mathbf p_e)
\times
(\mathbf p_e-\mathbf p_{\mathrm{sh}})
\right]
}{
\|\mathbf p_w-\mathbf p_e\|_2
\|\mathbf p_e-\mathbf p_{\mathrm{sh}}\|_2
},
\end{equation*}
with wrist, elbow, and shoulder positions $\mathbf p_w$, $\mathbf p_e$, and $\mathbf p_{\mathrm{sh}}$, and the robot elbow's world-frame bending axis $\mathbf n_s$.

\paragraph{Fixed-foot detection in the source.}
The sole-stick mask is inferred once from the unscaled source toe-base positions, ankle positions, and foot headings on the 50\,Hz time grid, before morphology adaptation. Toe height above the inferred support surface, ankle lift relative to its stable-contact height, horizontal and vertical toe speeds, and foot yaw rate determine continuous contact, support, stance, settling, motion, swing, liftoff, and landing scores in $[0,1]$, with causal smoothing. For ground support, the stationarity gate requires horizontal toe speed at most $0.15$\,m/s and absolute yaw rate at most $30^\circ$/s. Contact and transition scores additionally distinguish a planted foot from a slowly moving airborne foot.

The fixed-foot flag requires this stationarity gate, contact score at least $0.50$ or support score at least $0.55$, aggregate support strength at least $0.55$, and motion, swing, liftoff, and landing scores all below $0.25$. Aggregate support strength is the maximum of the support, stance, and settling scores and the release-attenuated contact score $c(1-r^3)$, where $c$ is the contact score and $r$ is the maximum motion, swing, and liftoff score. Only contiguous runs lasting at least $0.12$\,s (six frames at 50\,Hz) are retained. Each retained run receives a separate episode identifier, and the active feet define $\mathcal S_t$. The mask and identifiers remain fixed during optimization.

Support height is inferred per episode to include feet planted on raised surfaces. Candidate raised plateaus lie more than $12$\,cm above the global floor, persist for at least $0.12$\,s, and have a 5th--95th percentile height range at most $2.5$\,cm; their acceptance also checks descent onto the surface and subsequent departure. An accepted raised-support episode supplies the stationarity gate for the interaction datasets. LAFAN1 additionally requires horizontal and vertical toe speeds at most $0.30$ and $0.25$\,m/s and yaw rate at most $100^\circ$/s. The same contact, transition, and persistence conditions still apply.

\paragraph{Hard horizontal support constraints.}
At each frame, the body solver reads the fixed source mask to acquire, retain, or release a foot constraint. The constrained robot toe point $\mathbf c_s(\mathbf q)$ is located at $[0.14,0,-0.03]^\top$\,m in the G1 foot frame. At episode entry $t_e$, the solver stores its position $\mathbf c_{e,s}$ from the entry robot configuration and the corresponding mapped source contact pose $(\widehat R_{t_e,s},\widehat{\mathbf p}_{t_e,s})$. The first output frame establishes the initial anchors. During an episode, the desired horizontal anchor is
\begin{equation}
\mathbf a_{t,s}
=\Pi_{xy}\!\left[
\widehat{\mathbf p}_{t,s}
+\widehat R_{t,s}\widehat R_{t_e,s}^{\top}
\left(\mathbf c_{e,s}-\widehat{\mathbf p}_{t_e,s}\right)
\right],
\label{eq:app_support_anchor}
\end{equation}
where $\Pi_{xy}$ selects the horizontal coordinates. LAFAN1 uses its independently constructed source toe target with an episode-constant horizontal offset acquired from the entry configuration. Thus the constraint preserves the robot's contact offset relative to source support motion; a stationary source target gives a constant world-space anchor. The anchor is released when the source sole-stick flag becomes inactive.

For every $s\in\mathcal S_t$, the equality $\mathbf c_{s,xy}(\mathbf q)=\mathbf a_{t,s}$ enters both body IK passes as a hard QP constraint. At iterate $\mathbf q^{(k)}$, its linearization is
\begin{equation}
J_{s,xy}(\mathbf q^{(k)})\Delta\boldsymbol\xi
=\mathbf a_{t,s}-\mathbf c_{s,xy}(\mathbf q^{(k)}),
\label{eq:app_support_qp}
\end{equation}
where $\Delta\boldsymbol\xi$ is the tangent-space configuration increment. Joint and collision bounds are enforced in the same solve. Toe forward kinematics is checked again after integration; nonlinear step correction controls the equality residual with a nominal per-coordinate tolerance of $10^{-4}$\,m. Finite outputs that retain a larger residual are recorded as support-quality violations and remain subject to output-based skating evaluation. The equality constrains only horizontal toe position: height follows the support surface and nonpenetration constraints, while foot orientation is tracked through soft objectives. Subsequent hand fitting and upper-body refinement leave the floating base and legs fixed, preserving the attained support configuration.

\paragraph{Collision boundaries.}
The signed separation $d_c$ is positive outside collision. The set $\mathcal C_t^B$ includes ground and support-surface boundaries, filtered body self-collision pairs, and enabled body--object boundaries; detailed finger geometry is excluded. We use $\epsilon_c=0$, except for the GRAB and HUMOTO non-arm body--object check, which allows $\epsilon_c=0.02$\,m. Collision and support-equality tolerances are $10^{-5}$\,m and $10^{-4}$\,m, respectively. These optimization settings are distinct from evaluation thresholds.

\begin{table}[t]
\raggedright
\caption{
\textbf{Body-optimization objectives.}
Position and orientation-axis terms use squared Euclidean and
squared angular errors, respectively; posture terms act on joint
coordinates. Weights are per landmark, axis, or joint, with
$\mathrm{k}=10^3$.
Foot and knee weights are listed before support/contact modulation.
Lengths are in meters and angles in radians unless stated otherwise.
}
\label{tab:body_objectives}
\small
\setlength{\tabcolsep}{3pt}
\begin{tabular}{
@{}
p{0.24\linewidth}
p{0.50\linewidth}
p{\dimexpr0.26\linewidth-4\tabcolsep\relax}
@{}
}
\toprule
Term & Target & Weight or distance scale \\
\midrule

\multicolumn{3}{@{}l}{
\textit{Body and support tracking}
} \\

Pelvis pose
& World position; mapped $x/y/z$ axes.
& $82\mathrm{k}$;
  $(2.4,2.4,3.2)\mathrm{k}$ \\

Torso pose
& World position; mapped axes.
& $11\mathrm{k}$;
  $(1.05,1.05,1.45)\mathrm{k}$ \\

Shoulder / elbow positions
& World-frame shoulders; torso-relative elbows.
& $4.2\mathrm{k}$ ; $1.2\mathrm{k}$ \\

Wrist pose
& Interaction-dependent position; mapped axes.
& $240\mathrm{k}$;
  $(2.1,1.7,0.9)\mathrm{k}$ \\

Foot positions
& Support-aware sole center; heel and toe positions; sole-corner heights.
& $115\mathrm{k}$;
  $48\mathrm{k}$ each;
  $260\mathrm{k}$ each \\

Foot orientation
& Forward axis; sole normal.
& $1.8\mathrm{k}$; $2.8\mathrm{k}$ \\

Knee position
& World positions during ground-contact motion.
& $115\mathrm{k}$ \\

\midrule
\multicolumn{3}{@{}l}{
\textit{Temporal regularization ($t>1$)}
} \\

Root orientation
& Previous root correction transported by the source
  rotation change.
& $(1.2,1.2,1.6)\mathrm{k}$ \\

Torso / elbow corrections
& Previous pose/position corrections transported
  with the current reference.
& $w_{\mathrm{track}}\tau_B/\Delta t$ \\

Non-arm joint posture
& Previous leg-twist, other leg, waist, and remaining
  non-arm joint angles.
& $1.20$, $0.16$, $0.12$, $0.045$ \\

Arm joint posture
& Previous arm/wrist angles.
& $\tau_B h_j/(2\Delta t)$ \\

\midrule
\multicolumn{3}{@{}l}{
\textit{Posture and bending priors}
} \\

Leg bending
& Source bending direction/branch;
  optional knee-flexion targets outside ground episodes.
& $420o^2$; $420$ \\

Arm joint margins
& Neutral posture in joint-margin coordinates:
  wrist, shoulder/elbow, and other arm joints.
& $1.20$, $0.03$, $0.025$ \\

Initial body posture
& Neutral non-arm joint angles at the first frame.
& $0.018$ \\

\midrule
\multicolumn{3}{@{}l}{
\textit{Soft collision clearance}
} \\

Self / ground clearance
& Body geometry and arm proxies;
  self-clearance margin of 5\,mm and zero penetration
  into the ground or active support plane.
& 35\,mm \\

Object clearance
& Surface/material distance;
  zero margin at contact patches and 3\,mm elsewhere.
& 3\,mm for fine body geometry;
  35\,mm for coarse/arm geometry \\

Exact object penetration
& Robot collision convexes against object meshes;
  smooth zero-penetration penalty.
& 0.1\,mm \\

\bottomrule
\end{tabular}

\par\smallskip
\footnotesize
\rightskip=0pt\relax
Axis-weight triples follow $x,y,z$. Here $\tau_B=0.1$\,s, giving $\tau_B/\Delta t=5$ at 50\,Hz;
$w_{\mathrm{track}}$ is the corresponding tracking weight.
The quantity $h_j$ is the Gauss--Newton curvature of the arm-margin
term at the previous configuration, and $o\in[0,1]$ measures
source leg-bend observability.
Clearance residuals are based on $[m_c-d_c]_+$ and the listed
distance scales, where $[x]_+=\max(x,0)$.
Exact penetration uses a smoothed boundary transition.
Terms are enabled only when the required references or geometry
are available.
\end{table}

\subsubsection{Hand Codebook and Continuous Fitting}

Each hand is parameterized by six independent drivers. Dependent joints follow the mimic relation
$q_j=a_j\theta_{d(j)}+b_j$. Driver bounds are intersected with the bounds implied by all
dependent joints. Forward kinematics yields wrist-local descriptors of finger directions, accumulated bending angles, reach-to-chain-length ratios, and normalized inter-finger distances. Retrieval uses the mean squared distance over observed components selected by the validity mask $\mathbf m_t$:
\begin{equation}
\begin{aligned}
D_{tn}
&=
\frac{
\|\mathbf m_t\odot
(\mathbf g_t^{s,H}-\mathbf g_n^{s,R})\|_2^2
}{
\sum_j m_{t,j}
},
\\
\widehat{\boldsymbol\theta}_t^s
&=
\frac{
\sum_{n\in\mathcal I_t^s}
\exp(-D_{tn}/\tau)\boldsymbol\theta_n^s
}{
\sum_{n\in\mathcal I_t^s}
\exp(-D_{tn}/\tau)
}.
\end{aligned}
\label{eq:app_hand_retrieval}
\end{equation}
Here $\mathcal I_t^s$ contains the $K=12$ nearest entries and $\tau=0.06$.

Starting from $\widehat{\boldsymbol\theta}_t^s$, continuous fitting combines wrist-local fingertip and chain position residuals, a posture prior informed by the retrieved configurations and observed finger shape, and first- and second-order driver smoothness. Optimization updates only the independent drivers within their bounds, with dependent joints reconstructed through the mimic relations. Table~\ref{tab:hand_codebook_fitting} details the codebook construction, retrieval, and continuous-fitting settings. The fitted hands and body trajectory together form the decoupled initialization $\bar{\mathbf q}_{1:T}$. Sequences without articulated hand references use neutral fingers at this stage.

\begin{table}[t]
\raggedright
\caption{
\textbf{Hand-initialization settings.}
Fitting weights are per landmark or driver and multiply squared
position or driver-coordinate errors.
Position targets are wrist-local; lengths are in meters and
angles in radians.
Here $\mathrm{k}=10^3$.
}
\label{tab:hand_codebook_fitting}
\small
\setlength{\tabcolsep}{3pt}
\begin{tabular}{
@{}
p{0.24\linewidth}
p{0.46\linewidth}
p{\dimexpr0.30\linewidth-4\tabcolsep\relax}
@{}
}
\toprule
Term & Target and activation & Weight or setting \\
\midrule

\multicolumn{3}{@{}l}{
\textit{Codebook and retrieval}
} \\

Driver sampling
& Six mimic-feasible drivers: thumb yaw/pitch and four finger
  flexions; interval sampling followed by forward kinematics.
& $\{0.15,0.50,0.85\}$;
  $N_B=3^6=729$ per hand \\

Finger descriptors
& Wrist-to-tip, chain-root, and up to three segment unit vectors;
  accumulated bending and tip reach divided by chain length.
& Up to $15+2$ components per finger \\

Inter-finger distances
& Selected tip-to-tip distances normalized by the mean
  wrist-to-tip radius.
& $6$ pairs;
  at most $91$ descriptor components overall \\

Masked retrieval
& Observed-component distance and exponential driver blending
  in Eq.~\eqref{eq:app_hand_retrieval}.
& $K=12$; $\tau=0.06$ \\

\midrule
\multicolumn{3}{@{}l}{
\textit{Geometric fitting}
} \\

Fingertip positions
& Observed thumb, index, and remaining fingertips.
& $4.5\mathrm{k}\,f_{\mathrm{tip}}rp$ \\

Finger-chain positions
& Observed $j_1,j_2,j_3$, and fingertip landmarks.
& $2.25\mathrm{k}\,f_{\mathrm{chain}}\ell_jp$ \\

\midrule
\multicolumn{3}{@{}l}{
\textit{Posture and support priors}
} \\

Shape posture
& Six-driver reference informed by observed geometry
  and retrieved configurations.
& $170(1+3\sigma+6\eta)$ \\

Support hold
& Shape-reference driver angles during stable hand support.
& $374\,\alpha\mathbf c_{\mathrm{sup}}$ \\

Target hold
& Previous driver angles while holding an interaction target.
& $510\,\eta\mathbf c_{\mathrm{tgt}}$ \\

Thumb support
& Thumb yaw/pitch when a support pitch target is available.
& $1.2\mathrm{k}\,\beta(0.78,1.35)$ \\

\midrule
\multicolumn{3}{@{}l}{
\textit{Temporal regularization}
} \\

First-order smoothness
& Previous driver configuration, after the first frame.
& $460(1+4\sigma+10\eta)$ \\

Second-order smoothness
& Extrapolated driver configuration
  $2\boldsymbol\theta_{t-1}-\boldsymbol\theta_{t-2}$.
& $180(1+3\sigma+7\eta)$ \\

\bottomrule
\end{tabular}

\par\smallskip
\footnotesize
\rightskip=0pt\relax
Thumb/index/other-finger factors are $(1.45,1.25,0.78)$ for $f_{\mathrm{tip}}$ and $(1.25,1.08,0.82)$ for $f_{\mathrm{chain}}$. For $j_1,j_2,j_3$, and tip, respectively, $\ell_j$ takes $0.40$, $0.72$, $0.95$, and $0.85$; non-thumb $j_3$ landmarks receive an additional factor of $0.35$. The pass factor is $p=0.72$ for coarse fitting and $p=1.18$ for fine fitting; $r=1.45$ for the optional fingertip-focused solve and $r=1$ otherwise. The strengths $\sigma$, $\eta$, $\alpha$, and $\beta$ denote support stability, target hold, support hold, and thumb support, respectively. The per-driver factors $\mathbf c_{\mathrm{sup}}=(2,2,2.6,2.8,2.8,2.8)$ and $\mathbf c_{\mathrm{tgt}}=(1.8,2.1,2.8,3,3,3)$ follow the six-driver order. Coarse and fine IK use eight iterations each with damping $0.5$, subject to driver bounds and mimic relations. Unavailable observations and inactive support terms contribute no residual; fingers remain neutral without articulated hand observations.
\end{table}

\subsection{Coupled Upper-Body Interaction Refinement}
\label{sec:appendix_coupled}

Refinement jointly optimizes 14 arm/wrist joints and 12 independent hand drivers for OMOMO, GRAB, and HUMOTO. LAFAN1 updates only the arm/wrist joints and keeps the fingers neutral. The floating base, legs, and waist retain their initialized trajectories, and the preprocessed object trajectories remain unchanged.

\paragraph{Interaction objectives.}
For articulated-hand inputs, let $\mathbf p_{t,sf}$ and $\mathbf u_{t,sf}$ denote the world-frame and wrist-local positions of fingertip $f$ on hand $s$. Their corresponding targets are $\mathbf y_{t,sf}$ and $\mathbf v_{t,sf}$, and $\ell_{t,s}$ denotes the target hand scale. Using the interaction weight $\alpha_{t,s}$ from reference construction, fingertip alignment is defined as
\begin{equation}
\begin{aligned}
\mathcal L_{\mathrm{tip}}
=
\sum_{s,f}\frac{\omega_f}{\ell_{t,s}^2}
\Big[
&\alpha_{t,s}^2
\|\mathbf p_{t,sf}-\mathbf y_{t,sf}\|_2^2
+(1-\alpha_{t,s})^2
\|\mathbf u_{t,sf}-\mathbf v_{t,sf}\|_2^2
\Big].
\end{aligned}
\label{eq:app_tip_refinement}
\end{equation}
World-frame alignment preserves interaction placement, whereas wrist-local alignment preserves free-hand shape. Orientation terms use squared angles between corresponding palm-normal, wrist-forward, and lateral directions. Soft position and posture anchors in $\mathcal L_{\mathrm{other}}$ are listed in Table~\ref{tab:refinement_weights}. Wrist and palm position weights are additionally multiplied by $(1-\alpha_{t,s})^2$, relaxing their anchors as scene alignment becomes dominant.

\paragraph{Temporal regularization.}
Let $\boldsymbol\delta_t^U=\mathbf q_t^U-\bar{\mathbf q}_t^U$ denote the arm correction relative to the initialization. We regularize changes in this correction and in the hand drivers:
\begin{equation}
\begin{aligned}
\mathcal L_{\mathrm{temp}}
={}&
\lambda_U
\|D_U(\boldsymbol\delta_t^U-
       \boldsymbol\delta_{t-1}^U)\|_2^2
+
\lambda_H\sum_s
\|D_H(\boldsymbol\theta_t^s-
       \boldsymbol\theta_{t-1}^s)\|_2^2,
\end{aligned}
\label{eq:app_temporal}
\end{equation}
where $D_U$ and $D_H$ normalize each coordinate by its feasible range. The arm term preserves the previous correction relative to the current initialization rather than anchoring the arm to its previous pose. The hand term directly smooths driver motion.
Both terms are omitted at the first frame.

\begin{table}[t]
\raggedright
\caption{
\textbf{Coupled-refinement objectives.}
Position weights are divided by $\ell_{t,s}^2$ and joint-coordinate weights by squared feasible ranges. Activation factors are applied as described in the text. Angular residuals use radians.
}
\label{tab:refinement_weights}
\small
\setlength{\tabcolsep}{3pt}
\begin{tabular}{
@{}
p{0.24\linewidth}
p{0.46\linewidth}
p{\dimexpr0.30\linewidth-4\tabcolsep\relax}
@{}
}
\toprule
Term & Target and activation & Weight or setting \\
\midrule

\multicolumn{3}{@{}l}{
\textit{Interaction alignment}
} \\

Thumb / index fingertips
& World-frame and wrist-local targets in
  Eq.~\eqref{eq:app_tip_refinement}.
& $2.0$ each \\

Other fingertips
& Corresponding targets for the middle, ring, and little fingers.
& $0.20$ each \\

Hand orientation
& Palm normal, wrist-forward direction, and lateral axis.
& $0.55$, $0.42$, $0.12$ \\

\midrule
\multicolumn{3}{@{}l}{
\textit{Position and posture anchors}
} \\

Wrist / palm / elbow positions
& Soft position anchors; wrist and palm anchors relax
  with increasing interaction activation.
& $0.08$ / $0.06$ / $0.0125$ \\

Arm posture
& Posture prior in joint-limit-aware coordinates.
& $0.60$ \\

Hand posture
& Initialized hand-driver configuration.
& $0.035$ \\

\midrule
\multicolumn{3}{@{}l}{
\textit{Temporal regularization ($t>1$)}
} \\

Arm correction
& Previous correction relative to the current initialization.
& $\lambda_U=80.0$ \\

Hand drivers
& Previous independent driver configuration.
& $\lambda_H=0.40$ \\

\midrule
\multicolumn{3}{@{}l}{
\textit{Soft object clearance}
} \\

Hand--object clearance
& Clearance residual with an additional penalty inside material.
& $1.80$ \\

Noncontact margin
& Desired separation for noncontact samples.
& $3$\,mm \\

\bottomrule
\end{tabular}
\end{table}

\paragraph{Feasibility and collision handling.}
The refinement feasible set is
\begin{equation}
\mathcal Z_t=
\left\{
\mathbf z:
\begin{aligned}
&\mathbf q^-
\leq
\mathbf q_{\mathrm{act}}(\mathbf z;\bar{\mathbf q}_t)
\leq
\mathbf q^+,
\\
&q_j=a_j\theta_{d(j)}+b_j,
\\
&b_s\geq0
\quad(s\in\{L,R\}),
\\
&d_c\geq0
\quad(c\in\mathcal C_t^U)
\end{aligned}
\right\},
\label{eq:app_refinement_constraints}
\end{equation}
where $\mathbf q_{\mathrm{act}}(\mathbf z;\bar{\mathbf q}_t)$
contains the reconstructed actuated coordinates, with all inactive coordinates fixed to their initialized values. Mimic relations and elbow-bending constraints use the definitions from body--hand initialization. The set $\mathcal C_t^U$ contains active upper-limb self-collision boundaries and enabled ground constraints. HUMOTO and OMOMO enable ground checks for the full arm, palm, and finger geometry; LAFAN1 excludes palm and finger geometry for the reason of non-finger dataset. Adjacent and allowed robot pairs are filtered out, and fine self-collision uses URDF collision convexes. Ground and self-collision checks use a numerical tolerance of $10^{-5}$\,m.

Object clearance and contact recovery are soft objectives, rather than constraints in $\mathcal C_t^U$. For a sampled object distance $d_i$ and clearance margin $m_i$, the clearance residual is $r_i=[m_i-d_i]_+ + [-d_i]_+$, where $[x]_+=\max(x,0)$. The second term increases the penalty inside object material. Material penetration and visible-surface proximity use distinct geometric queries.

\subsection{Numerical Implementation}
\label{app:numerical_implementation}

Frames are processed in temporal order using damped differential IK
with Mink~\citep{zakka2026mink} and DAQP~\citep{arnstrom2022dual}.
The nominal coarse/fine iteration limits are 22/30 for body
optimization and 8/8 for hand fitting; refinement uses
40 iterations per solve.
Body damping is 0.30/0.25 for the coarse/fine passes,
and hand fitting and refinement use 0.50.
The body stage additionally uses constrained SLSQP
for arm optimization.
Feasibility recovery may require additional iterations.
After each integration step, nonlinear geometry is re-evaluated,
inactive coordinates remain fixed, and mimic relations are
enforced in both configuration and tangent coordinates.

\section{Policy Implementation and Training}
\label{app:policy_protocol}

Policies are trained in IsaacLab Simulator for a Unitree G1 equipped with Inspire RH56DFQ hands, with a separate policy for each retargeted reference. The network architecture is shown in Figure~\ref{fig:rl_policy}. The loco-manipulation comparison evaluates DexWeave and an object-conditioned MLP on different tasks. Within each reference, the actors share observations, action interfaces, rewards, privileged critics, randomization, and episode settings. 

\subsection{Observations and Network Configuration}

\paragraph{Actor observations.}
The loco-manipulation actor receives 238 inputs: 211 robot features and 27 object features, as detailed in Table~\ref{tab:actor_observations}. The torso serves as the reference anchor. Current joint positions are expressed relative to the default configuration, whereas reference joint positions remain absolute targets. Base angular velocity is expressed in the base frame, while anchor and object features use the current torso frame. The anchor position history contains seven samples spanning 120\,ms at 50\,Hz. Finger velocities are omitted from the actor input.

\begin{table}[t]
\raggedright
\caption{
\textbf{Actor and critic observations for dexterous loco-manipulation.}
Dimensions count scalar inputs before encoding.
The last column specifies regional assignments for actor inputs
and feature composition for critic inputs.
}
\label{tab:actor_observations}
\small
\setlength{\tabcolsep}{3pt}
\begin{tabular}{
@{}
>{\raggedright\arraybackslash}p{0.49\linewidth}
>{\centering\arraybackslash}p{0.12\linewidth}
>{\raggedright\arraybackslash}
p{\dimexpr0.39\linewidth-4\tabcolsep\relax}
@{}
}
\toprule
Observation & Dimension & Input details \\
\midrule

\multicolumn{3}{@{}l}{
\textit{Actor observations}
} \\

Reference body and hand joint positions
& 41 & Corresponding joint groups \\

Reference non-finger joint velocities
& 29 & Legs, waist, arms \\

Reference-anchor position history relative to torso
& $7\times3$ & Base \\

Reference-anchor orientation relative to torso
& 6 & Base \\

Base angular velocity
& 3 & Base \\

Current independent joint positions
& 41 & Corresponding joint groups \\

Current non-finger joint velocities
& 29 & Legs, waist, arms \\

Previous applied normalized action
& 41 & Corresponding joint groups \\

\cmidrule(lr){1-3}
Robot subtotal
& 211 & Eight regional tokens \\

Current and reference object poses and their coordinate difference
& $9+9+9$ & Object token \\

\textbf{Actor total}
& \textbf{238} & \\

\midrule
\multicolumn{3}{@{}l}{
\textit{Privileged critic observations (training only)}
} \\

Reference joint states
& 82 & 41 positions and 41 velocities \\

Anchor pose
& 9 & Position and orientation \\

Tracked-link poses
& 216 & 24 link positions and orientations \\

Base velocities
& 6 & Linear and angular velocities \\

Current joint states
& 82 & 41 positions and 41 velocities \\

Previous actions
& 41 & Previous action vector \\

Object features
& 27 & Current and reference poses and their coordinate difference \\

\cmidrule(lr){1-3}
\textbf{Critic total}
& \textbf{463} & Independent value-network input \\

\bottomrule
\end{tabular}

\par\smallskip
\footnotesize
Actor regional input dimensions are 30 for the base,
30 per leg, 15 for the waist, 35 per arm, and 18 per hand.
The actor omits finger velocities, whereas the critic includes
velocities for all 41 independent joint coordinates.
\end{table}

\paragraph{Object observations.}
For a pose $T=(R,\mathbf p)$, define $r_6(R)=[R_{11},R_{12},R_{21},R_{22},R_{31},R_{32}]^\top$ and $\phi(T)=[\mathbf p^\top,r_6(R)^\top]^\top$. The object observation is
\begin{equation}
\mathbf x_o=
\begin{bmatrix}
\phi(T_o)\\
\phi(T_o^{\mathrm{ref}})\\
\phi(T_o)-\phi(T_o^{\mathrm{ref}})
\end{bmatrix}
\in\mathbb R^{27}.
\end{equation}
Both poses are expressed in the same measured torso frame. The final nine components are coordinate-wise differences of the pose representations, rather than a relative rigid-body transform. The reference object trajectory remains prescribed in the scene and the simulated object moves dynamically.

\paragraph{Privileged critic observations.}
During training, the critic receives privileged observations, including finger velocities, tracked-link poses, and base linear velocity that are unavailable to the actor. Robot and object states are obtained directly from simulation without the observation noise or delay applied to the actor. The critic inputs are concatenated and processed by an independent value network. Table~\ref{tab:actor_observations} lists the complete critic observations.

\paragraph{Network configuration.}
Each regional encoder has one 128-unit hidden layer and produces a 128-dimensional token. The object encoder receives the separate 27-dimensional object observation. The robot Transformer contains two blocks and the object-fusion Transformer contains one. Each block uses four attention heads, a feed-forward width of 256, residual connections, layer normalization, ELU activations, and no dropout. Each action head has one 128-unit hidden layer. The independent value network has hidden widths $(512,256,128)$. Empirical observation normalization is applied to network inputs. The object-conditioned MLP baseline has three hidden layers of width 621 and uses the same critic.

\paragraph{Auxiliary motion tracking task.}
The body-only actor uses 157 inputs, six body tokens, and 29 actions. It retains one anchor-position sample and omits hand observations, hand actions, and object conditioning. Its privileged critic receives 286 inputs. The attention-mask ablations change only the permitted connections described in Sec.~\ref{sec:ablation}.

\subsection{Action Interface and Reward Design}

\paragraph{Joint-position control.}
The policy outputs 41 normalized commands: 12 for the legs, three for the waist, 14 for the arms and wrists, and 12 for the independent hand drivers. These commands define joint position targets as
\begin{equation}
\mathbf q_t^{\mathrm{cmd}}
=
\mathbf q^{\mathrm{default}}
+
\mathbf s\odot\mathbf a_t,
\end{equation}
where $s_j=0.25\tau_j^{\max}/k_{p,j}$ for body joints and $s_j=0.25$ for hand drivers. Joint-level PD controllers execute the targets, with nominal hand-load gravity compensation applied to the waist and arms. Finger gains are $k_p=20$\,N\,m/rad and $k_d=1$\,N\,m\,s/rad, with torque and velocity limits of 2\,N\,m and 5\,rad/s. Leg and waist target velocities are limited to 16\,rad/s, except for hip yaw at 20\,rad/s. The previous-action observation records the normalized target after this command processing.

The 12 independent hand commands are expanded through fixed mimic relations into 24 articulated finger joints, yielding 53 simulated joints in total. The thumb's intermediate and distal joints use ratios of 1.6 and 2.4 relative to the pitch driver; the other four fingers use unit-ratio coupling. Reference observations and action decoding use consistent anatomical joint groups.

\paragraph{Reward terms.}
Tracking rewards take the form $r_k=\exp(-E_k/\sigma_k^2)$, where $E_k$ is a squared position, joint-angle, velocity, or rotation-angle error. Orientation errors use geodesic angles on $\mathrm{SO}(3)$. For multi-link and joint terms, squared errors are averaged over the selected elements before applying the exponential. Body-pose rewards use the aligned reference defined in Appendix~\ref{app:evaluation}. Hand poses are evaluated relative to their respective wrists, and wrist--object terms compare relative poses. Table~\ref{tab:policy_rewards} gives the tracking weights, error scales, and control penalties.

\begin{table}[t]
\raggedright
\caption{
\textbf{Reward configuration for dexterous loco-manipulation.}
Tracking rewards use the listed weights and error scales $\sigma$, and penalty terms use direct coefficients. Lengths are in meters and angles in radians.
}
\label{tab:policy_rewards}
\small
\setlength{\tabcolsep}{3pt}
\begin{tabular}{
@{}
>{\raggedright\arraybackslash}p{0.58\linewidth}
>{\centering\arraybackslash}p{0.20\linewidth}
>{\centering\arraybackslash}
p{\dimexpr0.22\linewidth-4\tabcolsep\relax}
@{}
}
\toprule
Term & Weight & Scale $\sigma$ \\
\midrule

\multicolumn{3}{@{}l}{\textit{Whole-body tracking}} \\

Global torso position / orientation
& $1.5 / 1.5$ & $0.20 / 0.20$ \\

Aligned body position / orientation
& $1.0 / 1.0$ & $0.30 / 0.40$ \\

Global body linear / angular velocity
& $1.0 / 1.0$ & $1.0 / 3.14$ \\

Body joint positions
& $1.0$ & $0.40$ \\

\midrule
\multicolumn{3}{@{}l}{\textit{Hand and object interaction}} \\

Independent hand-driver positions
& $0.3$ & $0.50$ \\

Wrist-relative hand position / orientation
& $0.3 / 0.3$ & $0.10 / 0.40$ \\

Wrist--object relative position / orientation
& $1.2 / 1.0$ & $0.20 / 0.40$ \\

Object position / orientation
& $1.2 / 1.0$ & $0.15 / 0.90$ \\

\midrule
\multicolumn{3}{@{}l}{\textit{Control regularization}} \\

Action first difference, squared norm
& $-0.10$ & -- \\

Policy-output second difference, waist / legs
& $-0.05 / -0.01$ & -- \\

Joint-limit violation
& $-10.0$ & -- \\

Undesired contacts above 1\,N
& $-0.10$ & -- \\

\bottomrule
\end{tabular}
\end{table}

The undesired-contact penalty excludes permitted foot and hand contacts. Second-order action penalties are computed from policy outputs before command processing. For body-only tracking, we omit hand and object rewardsand the second-order leg-action penalty. The torso position and orientation rewards use weights of $1.0$ and $0.75$, with corresponding error scales of $0.20$\,m and $0.30$\,rad.

\subsection{Policy Optimization and Domain Randomization}

\paragraph{PPO training.}
All actor components and the privileged critic are optimized jointly with PPO using the settings in Table~\ref{tab:policy_protocol}. Each update partitions the collected transitions into four minibatches of 24,576 samples. The learning rate is adapted using the target KL divergence, and the value loss uses clipping. Loco-manipulation metrics are aggregated over the common training window specified in Appendix~\ref{app:evaluation}.

\begin{table}[t]
\raggedright
\caption{\textbf{PPO and simulation settings.}}
\label{tab:policy_protocol}
\small
\setlength{\tabcolsep}{3pt}
\begin{tabular}{
@{}
>{\raggedright\arraybackslash}p{0.28\linewidth}
>{\raggedright\arraybackslash}p{0.17\linewidth}
>{\raggedright\arraybackslash}p{0.34\linewidth}
>{\raggedright\arraybackslash}
p{\dimexpr0.21\linewidth-6\tabcolsep\relax}
@{}
}
\toprule
Setting & Value & Setting & Value \\
\midrule

Parallel environments & 4,096
& Physics / control frequency & 200 / 50\,Hz \\

Steps per environment & 24
& Samples per update & 98,304 \\

Learning epochs / minibatches & 5 / 4
& Initial learning rate & $10^{-3}$ \\

PPO clipping parameter & 0.2
& Target KL divergence & 0.01 \\

Discount $\gamma$ / GAE $\lambda$ & 0.99 / 0.95
& Value-loss coefficient & 1.0 \\

Maximum gradient norm & 1.0
& Initial action standard deviation & 1.0 \\

Entropy coefficient, tracking / manipulation & 0.005 / 0.002 \\

\bottomrule
\end{tabular}
\end{table}

\paragraph{Dynamics and observations domain randomization.}
During training, we randomize contact and inertial properties, actuation and control parameters, and actuator delays. External disturbances are applied as velocity perturbations at randomized intervals. Actor observations additionally include measurement noise and delays, with object poses sampled at a fixed frequency. The critic receives the corresponding states from the same randomized simulation without the observation noise or delays applied to the actor. Table~\ref{tab:policy_randomization} summarizes the randomization models and ranges, together with the fixed measurement settings.

\begin{table}[t]
\raggedright
\caption{
\textbf{Domain randomization and measurement settings.}
Randomized quantities are sampled uniformly over the listed ranges;
$\pm a$ denotes $[-a,a]$.
Multiplicative factors are relative to nominal values.
Fixed settings are marked explicitly.
}
\label{tab:policy_randomization}
\small
\setlength{\tabcolsep}{3pt}
\begin{tabular}{
@{}
>{\raggedright\arraybackslash}p{0.34\linewidth}
>{\raggedright\arraybackslash}p{0.23\linewidth}
>{\raggedright\arraybackslash}
p{\dimexpr0.43\linewidth-4\tabcolsep\relax}
@{}
}
\toprule
Quantity & Model & Range or value \\
\midrule

\multicolumn{3}{@{}l}{
\textit{Contact and inertial properties}
} \\

Static friction
& Coefficient
& $[0.4,1.8]$ \\

Dynamic friction
& Coefficient
& $[0.35,1.5]$ \\

Restitution
& Fixed coefficient
& $0$ \\

Foot contact offset
& Distance parameter
& $[5,20]$\,mm \\

Body mass
& Multiplicative
& $[0.9,1.1]$ \\

Hand payload
& Multiplicative
& $[0.8,1.25]$ \\

Torso center of mass
& Position offset
& $\pm(0.025,0.05,0.05)$\,m \\

Hand center of mass
& Position offset
& $\pm0.03$\,m per axis \\

\midrule
\multicolumn{3}{@{}l}{
\textit{Actuation and control}
} \\

Motor strength
& Multiplicative
& $[0.8,1.2]$ \\

Joint armature
& Multiplicative
& $[0.8,1.2]$ \\

PD stiffness
& Multiplicative
& $[0.85,1.15]$ \\

PD damping
& Multiplicative
& $[0.75,1.25]$ \\

Default joint position
& Additive offset
& $\pm0.02$\,rad \\

Joint friction
& Parameter value
& $[0,0.08]$ \\

Actuator delay
& Physics-step delay
& 0--2 steps (0--10\,ms) \\

\midrule
\multicolumn{3}{@{}l}{
\textit{External disturbances}
} \\

Push interval
& Event interval
& $[1.5,3.0]$\,s \\

Horizontal velocity
& Velocity disturbance
& $\pm0.4$\,m/s in $x$ and $y$ \\

Roll/pitch angular velocity
& Velocity disturbance
& $\pm0.15$\,rad/s \\

Yaw angular velocity
& Velocity disturbance
& $\pm0.25$\,rad/s \\

\midrule
\multicolumn{3}{@{}l}{
\textit{Actor observation noise}
} \\

Joint position
& Additive noise
& $\pm0.01$\,rad \\

Joint velocity
& Additive noise
& $\pm0.5$\,rad/s \\

Base angular velocity
& Additive noise
& $\pm0.2$\,rad/s \\

Anchor position
& Additive noise
& $\pm0.05$\,m \\

Anchor orientation
& Representation noise
& $\pm0.05$ per component \\

Object position
& Translation perturbation
& $\pm0.01$\,m \\

Object orientation
& Rotation perturbation
& $\pm0.035$\,rad \\

\midrule
\multicolumn{3}{@{}l}{
\textit{Actor measurement timing}
} \\

Anchor-position delay
& Measurement delay
& $[0,120]$\,ms \\

Object measurement rate
& Fixed sampling rate
& $30$\,Hz \\

Object measurement latency
& Measurement delay
& $[20,60]$\,ms \\

\bottomrule
\end{tabular}

\par\smallskip
\footnotesize
Static and dynamic friction use consistent coefficients.
Torso center-of-mass offsets follow the $x,y,z$ axes.
Anchor orientation noise acts on rotation-representation components,
whereas object orientation perturbations are specified in radians.
The critic receives simulation observations without the noise
or measurement delays applied to the actor.
\end{table}

\subsection{Episode Initialization and Termination}

\paragraph{Reference-phase initialization.}
The default initialization scheme uses equal proportions of first-frame and sampled-phase starts. Robot and object poses are initialized at the same reference phase, and their velocities follow the reference without added impulses. For sampled starts, phases are drawn using 70\% uniform sampling and 30\% failure-adaptive sampling. The adaptive weights use the square root of phase-wise failure rates, with phase-bin probabilities capped at 0.28. Sampled-start episodes use 0.5--1.5\,s of pre-roll and last at most 10\,s.

\paragraph{Termination conditions.}
Episodes end at the reference end or the applicable time limit, or earlier when a failure condition is met. Both loco-manipulation and body-only tracking terminate when the torso or end-effector height error exceeds 0.25\,m. For loco-manipulation, additional failure thresholds are 0.50\,m for torso horizontal position error and 0.70\,rad for torso orientation error. Object position and orientation errors are limited to 0.30\,m and 2.0\,rad, respectively. Body-only tracking additionally terminates when the vertical projected-gravity discrepancy exceeds 0.8 and has a time limit of 500 control steps.
\section{Evaluation Definitions}\label{app:evaluation}

\subsection{Retargeting Metrics}

\paragraph{Evaluation setup.}
Each method is evaluated using its corresponding robot and hand geometry and physical sampling interval. Source contact and support schedules are temporally aligned with the output frames. For the modular baselines, DexPilot or SBR initializes finger articulation on the OmniRetarget body trajectory. The additional IK variants optimize the two seven-joint arm--wrist chains using world-frame wrist and fingertip position objectives, with the fingers, floating base, waist, and legs fixed. The refinement ablation compares DexWeave's initialized and refined outputs on matched sequences, verifying that all inactive body coordinates and object trajectories remain identical.

\paragraph{Penetration.}
Penetration is recomputed from the saved output configuration at every frame. For each applicable component $c\in\{\mathrm{ground},\mathrm{self},\mathrm{object}\}$, let $p_{t,c}\geq0$ be the deepest detected penetration at frame $t$. Ground and filtered self-penetration use each method's MuJoCo G1 collision geometry, excluding structurally allowed pairs, while palm and finger geometry are excluded for LAFAN1. OMOMO evaluates robot collision-surface points against its material signed distance field, whereas GRAB and HUMOTO compare robot URDF collision convexes against the original object triangle meshes. The component thresholds are $\tau_{\mathrm{ground}}=\tau_{\mathrm{self}}=1$\,cm and $\tau_{\mathrm{object}}=2$\,cm for all interaction datasets. LAFAN1 has no object component. For a sequence of $T$ output frames,
\begin{equation}
D_{\mathrm{pen}}
=\frac{1}{T}\sum_{t=1}^{T}
\mathbf{1}\!\left[\exists c:\ p_{t,c}>\tau_c\right].
\label{eq:app_penetration_duration}
\end{equation}
Here $\mathbf{1}[\cdot]$ is the indicator function. A frame is counted once even if several components violate their thresholds. Raw penetration depths are retained without subtracting the thresholds. These evaluation thresholds are applied to the returned trajectory independently of the optimizer's clearance objectives, feasibility tolerances, or convergence status.

\paragraph{Foot skating.}
The fixed-foot mask and episode identifiers are derived from source motion as detailed in Appendix~\ref{app:body_support}, then sampled at each method's output times. They are independent of the optimized robot toe speed. At $t>1$, let $\mathcal V_t$ contain the feet whose source sole-stick flag is active at both $t-1$ and $t$ with the same episode identifier. The first frame and transitions into a new episode provide no valid within-episode displacement for that foot. Using the robot toe point defined in Appendix~\ref{app:body_support}, its horizontal speed is
\begin{equation}
v_{t,s}
=\frac{\|\mathbf c_{s,xy}(\mathbf q_t)
-\mathbf c_{s,xy}(\mathbf q_{t-1})\|_2}{\Delta t},
\end{equation}
where $\Delta t$ is the evaluated output's physical sampling interval. A frame is classified as skating when any $s\in\mathcal V_t$ has $v_{t,s}>0.30$\,m/s. The sequence-level duration is
\begin{equation}
D_{\mathrm{skate}}
=\frac{\displaystyle\sum_{t=2}^{T}
\mathbf{1}\!\left[\exists s\in\mathcal V_t:
v_{t,s}>0.30\,\mathrm{m/s}\right]}
{\displaystyle\sum_{t=2}^{T}
\mathbf{1}[\mathcal V_t\ne\emptyset]}.
\label{eq:app_skating_duration}
\end{equation}
The denominator counts frames with at least one eligible support foot, counting double support once. Sequences with a zero denominator have undefined skating metrics and are excluded from skating averages. The $0.30$\,m/s threshold classifies output violations; source support activation and hard optimization constraints use the separate rules in Appendix~\ref{app:body_support}.

\paragraph{Contact preservation and hand alignment.}
Evaluation frames are determined by source interactions, independently of robot contact. OMOMO detects contact using unsigned distance threshold to the visible object surface. Source detection uses the minimum distance over five reconstructed distal-finger proxies per hand and robot detection uses samples of its native hand collision surface. GRAB and HUMOTO instead select each hand's interaction frames using positive source finger-contact labels.

On OMOMO, contact preservation is the fraction of source-contact frames in which either robot hand contacts any active object part, without requiring the same hand or part as in the source. Contact distance averages unsigned distances from all five robot fingertips to the nearest active object surface on frames when the corresponding source hand is in contact. SOMA uses five fixed surface proxies on its rigid rubber hand. For finer alignment on GRAB and HUMOTO, primary and secondary errors measure world-frame fingertip distances for the thumb, index and remaining fingers, respectively. Palm error measures the angle between semantic palm normals. Targets come from the source scene before morphology blending or optimization adjustments.

\subsection{Policy Metrics and Aggregation}
\paragraph{Geometric errors.}
Let $a$ denote the torso anchor and $b$ a tracked link. Body-position error uses the yaw-aligned, height-preserving reference transform of BeyondMimic~\citep{liao2025beyondmimic}:
\begin{equation}
\begin{aligned}
R_{\Delta,t}
&=
R_z\!\left(
\operatorname{yaw}\!\left(
R_{t,a}(R^{\mathrm{ref}}_{t,a})^\top
\right)
\right),
\\
\widetilde{\mathbf p}^{\mathrm{ref}}_{t,b}
&=
\begin{bmatrix}
p_{t,a,x}\\
p_{t,a,y}\\
p^{\mathrm{ref}}_{t,a,z}
\end{bmatrix}
+
R_{\Delta,t}
\left(
\mathbf p^{\mathrm{ref}}_{t,b}
-
\mathbf p^{\mathrm{ref}}_{t,a}
\right).
\end{aligned}
\label{eq:app_reference_alignment}
\end{equation}
This transform aligns the reference yaw and horizontal anchor position with the robot while preserving the reference height. The body-position error at frame $t$ is
\begin{equation}
e_{\mathrm{body},t}
=
\frac{1}{|\mathcal B|}
\sum_{b\in\mathcal B}
\left\|
\mathbf p_{t,b}
-
\widetilde{\mathbf p}^{\mathrm{ref}}_{t,b}
\right\|_2.
\end{equation}
For body-only tracking, $\mathcal B$ contains 14 links: the pelvis, torso, and bilateral hip-roll, knee, ankle-roll, shoulder-roll, elbow, and wrist-yaw links. Loco-manipulation additionally includes ten hand landmarks.

Anchor and object errors are evaluated directly in the world frame, without reference alignment. Position error is the Euclidean distance between corresponding positions, and orientation error is
\begin{equation}
d_{\mathrm{SO}(3)}(R_1,R_2)
=
\arccos\!\left(
\operatorname{clip}\!\left(
\frac{\operatorname{tr}(R_1^\top R_2)-1}{2},
-1,1
\right)
\right).
\end{equation}

\paragraph{Episode statistics.}
Episode length is measured in control steps, with a maximum of 500 steps for body-only tracking. An episode is considered complete if it reaches the reference end or the prescribed time limit without early failure termination. Completion ratio for body-only tracking and success rate for loco-manipulation report the percentage of completed evaluation episodes. For sampled-phase starts, completion refers to the evaluated segment rather than necessarily the entire reference.

\paragraph{Evaluation protocol and aggregation.}
Body-only tracking and loco-manipulation policies are trained for 8k and 12k PPO iterations, respectively. During the final 1k iterations, policy checkpoints are saved every 100 iterations. For each reference sequence, every saved checkpoint is evaluated in MuJoCo five times using different evaluation seeds, with policy weights fixed throughout evaluation. All tabulated policy results are computed from these evaluation runs rather than training logs. For each metric, we first average across the five evaluation runs for each checkpoint, then across checkpoints within each sequence, and finally equally across reference sequences within each task setting.

\begin{table}
    \centering
    \caption{
        \textbf{Complete fine-grained interaction retargeting results on GRAB and HUMOTO.}
        Penetration duration is the fraction of colliding frames. Primary and secondary errors measure thumb/index and remaining-finger positions; palm error measures normal alignment. Lower is better. \bestvalue{Yellow} and \underline{underlined} mark the best and second-best values within each dataset.
    }
    \label{tab:retarget_dexterous_full}

    \small
    \setlength{\tabcolsep}{4pt}

    \begin{tabular*}{\linewidth}{@{\extracolsep{\fill}}llccccc}
        \toprule
        & & \multicolumn{2}{c}{\textbf{Penetration}}
        & \multicolumn{3}{c}{\textbf{Hand Alignment}} \\

        \cmidrule(lr){3-4}
        \cmidrule(lr){5-7}

        \textbf{Dataset} & \textbf{Method}
        & \shortstack{Duration \\ $\downarrow$}
        & \shortstack{Max Depth \\ (cm) $\downarrow$}
        & \shortstack{Pri. Err. \\ (mm) $\downarrow$}
        & \shortstack{Sec. Err. \\ (mm) $\downarrow$}
        & \shortstack{Palm Err. \\ (°) $\downarrow$} \\

        \midrule
        \multirow{5}{*}{\textit{GRAB}} & OmniRetarget + DexPilot
        & $\underline{0.183}$ & 2.123 & 209.353 & 207.745 & 52.122 \\

        & OmniRetarget + DexPilot \textbf{+ IK}
        & 0.212 & 2.269 & 16.585 & \bestvalue{$13.368$} & 11.555 \\

        & OmniRetarget + SBR
        & 0.184 & $\underline{2.120}$ & 210.524 & 212.985 & 52.122 \\

        & OmniRetarget + SBR \textbf{+ IK}
        & 0.190 & 2.362 & $\underline{14.925}$ & 14.659 & $\underline{9.530}$ \\

        & DexWeave (Ours)
        & \bestvalue{$0.001$} & \bestvalue{$1.263$} & \bestvalue{$4.342$} & $\underline{14.476}$ & \bestvalue{$3.652$} \\

        \midrule
        \multirow{5}{*}{\textit{HUMOTO}} & OmniRetarget + DexPilot
        & $\underline{0.370}$ & 3.638 & 292.886 & 278.574 & 47.815 \\

        & OmniRetarget + DexPilot \textbf{+ IK}
        & 0.522 & 3.597 & 32.389 & 31.018 & 14.444 \\

        & OmniRetarget + SBR
        & $\underline{0.370}$ & 3.638 & 292.750 & 283.381 & 47.815 \\

        & OmniRetarget + SBR \textbf{+ IK}
        & 0.511 & $\underline{3.582}$ & $\underline{29.999}$ & $\underline{28.436}$ & $\underline{12.476}$ \\

        & DexWeave (Ours)
        & \bestvalue{$0.007$} & \bestvalue{$2.854$} & \bestvalue{$7.198$} & \bestvalue{$11.702$} & \bestvalue{$4.439$} \\

        \bottomrule
    \end{tabular*}
\end{table}

\section{Additional Retargeting Results}\label{app:retargeting_results}

Table~\ref{tab:retarget_dexterous_full} gives the full comparison, including baselines with and without arm\textendash{}wrist IK. Adding IK substantially lowers fingertip error but increases penetration duration on both datasets. DexWeave improves primary-fingertip alignment and reduces penetration duration relative to both variants.


\section{Ablation Study}\label{sec:ablation}

\paragraph{Effect of upper-limb refinement.}

We compare the decoupled initialization and refined output on matched GRAB and HUMOTO sequences to isolate the contribution of coupled refinement (Table~\ref{tab:retarget_ablation}). Refinement reduces primary-fingertip error by 87.9\% on GRAB and 78.0\% on HUMOTO, together with lower secondary-fingertip errors. Penetration duration decreases from 0.157 to 0.001 and from 0.204 to 0.007, respectively. These improvements are accompanied by larger palm-normal errors on both datasets. On HUMOTO, Max Depth also increases from 2.568 to 2.854\,cm. Because this metric averages frame-wise maximum depths only over violating frames, the increase reflects the severity of the remaining violations rather than their frequency. Overall, refinement improves fingertip alignment and reduces penetration frequency, with a trade-off in palm orientation and conditional penetration depth on HUMOTO.

\begin{table}
    \centering
    \caption{\textbf{Retargeting-stage ablation.} Decoupled is the body-and-hand initialization; refined is the final output.}
    \label{tab:retarget_ablation}
    \small
    \setlength{\tabcolsep}{4pt}
    \begin{tabular*}{\linewidth}{@{\extracolsep{\fill}}llccccc}
        \toprule
        & & \multicolumn{2}{c}{\textbf{Penetration}} & \multicolumn{3}{c}{\textbf{Hand Alignment}} \\
        \cmidrule(lr){3-4}\cmidrule(lr){5-7}
        \textbf{Dataset} & \textbf{Stage} & \shortstack{Duration\\$\downarrow$} & \shortstack{Max Depth\\(cm) $\downarrow$} & \shortstack{Pri. Err.\\(mm) $\downarrow$} & \shortstack{Sec. Err.\\(mm) $\downarrow$} & \shortstack{Palm Err.\\(deg) $\downarrow$} \\
        \midrule
        \multirow{2}{*}{\textit{GRAB}} & Decoupled & 0.157 & 1.778 & 36.033 & 36.995 & 0.189 \\
        & Refined & 0.001 & 1.263 & 4.342 & 14.476 & 3.652 \\
        \midrule
        \multirow{2}{*}{\textit{HUMOTO}} & Decoupled & 0.204 & 2.568 & 32.703 & 26.090 & 0.159 \\
        & Refined & 0.007 & 2.854 & 7.198 & 11.702 & 4.439 \\
        \bottomrule
    \end{tabular*}
\end{table}

\begin{table}
    \centering
    \caption{\textbf{Policy-mask ablation.} Results are averaged over two reference motions using checkpoints saved every 100 iterations during the final 1,000 training iterations, with five evaluation seeds per checkpoint and reference. Dense $M_R$ and bidirectional $M_O$ each relax one attention mask. }
    \label{tab:mask_ablation}
    \small
    \setlength{\tabcolsep}{4pt}
    \begin{tabular*}{\linewidth}{@{\extracolsep{\fill}}lcccc}
        \toprule
        \textbf{Attention setting} & \shortstack{Complete Ratio\\(\%) $\uparrow$} & \shortstack{Body pos.\\(cm) $\downarrow$} & \shortstack{Object pos.\\(cm) $\downarrow$} & \shortstack{Object rot.\\(deg) $\downarrow$} \\
        \midrule
        $M_R$ dense & 91.30 & 4.79 & \textbf{3.62} & \textbf{8.23} \\
        $M_O$ bidirectional & 99.71 & 4.63 & 3.88 & 9.19 \\
        $M_R$ and $M_O$ (DexWeave) & \textbf{100} & \textbf{4.58} & 3.81 & 8.47 \\
        \bottomrule
    \end{tabular*}
\end{table}

\paragraph{Effect of the attention masks.}
We relax each attention mask separately, keeping the remaining architecture and training settings fixed. Dense $M_R$ enables unrestricted attention among all eight robot-region tokens. Bidirectional $M_O$ enables dense attention among the object token and five upper-body tokens, retaining the lower-body bypass. Table~\ref{tab:mask_ablation} shows that the full design achieves the highest completion ratio of 100\% and the lowest body-position error of 4.58\,cm. Dense $M_R$ yields lower object-position and orientation errors (3.62\,cm and 8.23$^\circ$, compared with 3.81\,cm and 8.47$^\circ$ for DexWeave), but its completion ratio falls to 91.30\%. Bidirectional $M_O$ achieves a similar completion ratio of 99.71\%, with higher body-position and object-pose errors than the full design. These results support directed attention for episode completion and body tracking, without showing a uniform advantage in object-pose accuracy.

\end{document}